\documentclass[letterpaper]{article} % DO NOT CHANGE THIS
\usepackage{aaai25}  % DO NOT CHANGE THIS
\usepackage{times}  % DO NOT CHANGE THIS
\usepackage{helvet}  % DO NOT CHANGE THIS
\usepackage{courier}  % DO NOT CHANGE THIS
\usepackage[hyphens]{url}  % DO NOT CHANGE THIS
\usepackage{graphicx} % DO NOT CHANGE THIS
\usepackage{natbib}  % DO NOT CHANGE THIS AND DO NOT ADD ANY OPTIONS TO IT
\usepackage{caption} % DO NOT CHANGE THIS AND DO NOT ADD ANY OPTIONS TO IT
\usepackage{subcaption} % Added for subfigure environment
\usepackage{algorithm}
\usepackage{algorithmic}

\usepackage{newfloat}
\usepackage{listings}
\DeclareCaptionStyle{ruled}{labelfont=normalfont,labelsep=colon,strut=off} % DO NOT CHANGE THIS
\floatstyle{ruled}
\newfloat{listing}{tb}{lst}{}
\floatname{listing}{Listing}
\usepackage{amsmath}
\usepackage{amsfonts}
\usepackage{amssymb} 
\usepackage{booktabs}
\usepackage{multirow}

\title{FilterTS: Comprehensive Frequency Filtering for Multivariate Time Series Forecasting}
\author {
    Yulong Wang\textsuperscript{\rm 1,\rm 2},
    Yushuo Liu\textsuperscript{\rm 1,\rm 2},
    Xiaoyi Duan\textsuperscript{\rm 1,\rm 2},
    Kai Wang\textsuperscript{\rm 1,\rm 2,\thanks{Corresponding author}}
}
\affiliations {
    \textsuperscript{\rm 1}College of Computer Science, Nankai University\\
    \textsuperscript{\rm 2}Tianjin Media Computing Center\\
    \{yl.wang, yushuo.liu, xyd\}@mail.nankai.edu.cn, wangk@nankai.edu.cn
}

\usepackage{bibentry}
\begin{document}

\maketitle
\begin{abstract}
Multivariate time series forecasting is crucial across various industries, where accurate extraction of complex periodic and trend components can significantly enhance prediction performance. However, existing models often struggle to capture these intricate patterns.
To address these challenges, we propose FilterTS, a novel forecasting model that utilizes specialized filtering techniques based on the frequency domain. 
FilterTS introduces a Dynamic Cross-Variable Filtering Module, a key innovation that dynamically leverages other variables as filters to extract and reinforce shared variable frequency components across variables in multivariate time series. Additionally, a Static Global Filtering Module captures stable frequency components, identified throughout the entire training set.
Moreover, the model is built in the frequency domain, converting time-domain convolutions into frequency-domain multiplicative operations to enhance computational efficiency.
Extensive experimental results on eight real-world datasets have demonstrated that FilterTS significantly outperforms existing methods in terms of prediction accuracy and computational efficiency.
\end{abstract}
\begin{links}
\link{code}{https://github.com/wyl010607/FilterTS}
\end{links}


\section{Introduction}

Multivariate time series forecasting is crucial across various domains such as finance\cite{olivares2023neural}, transportation\cite{bui2022spatial}, and energy\cite{zhou2021informer}, where the data often exhibits complex temporal characteristics like periodic changes and trends\cite{TFB}.

\begin{figure}[t!]
\centering
\includegraphics[width=1\columnwidth]{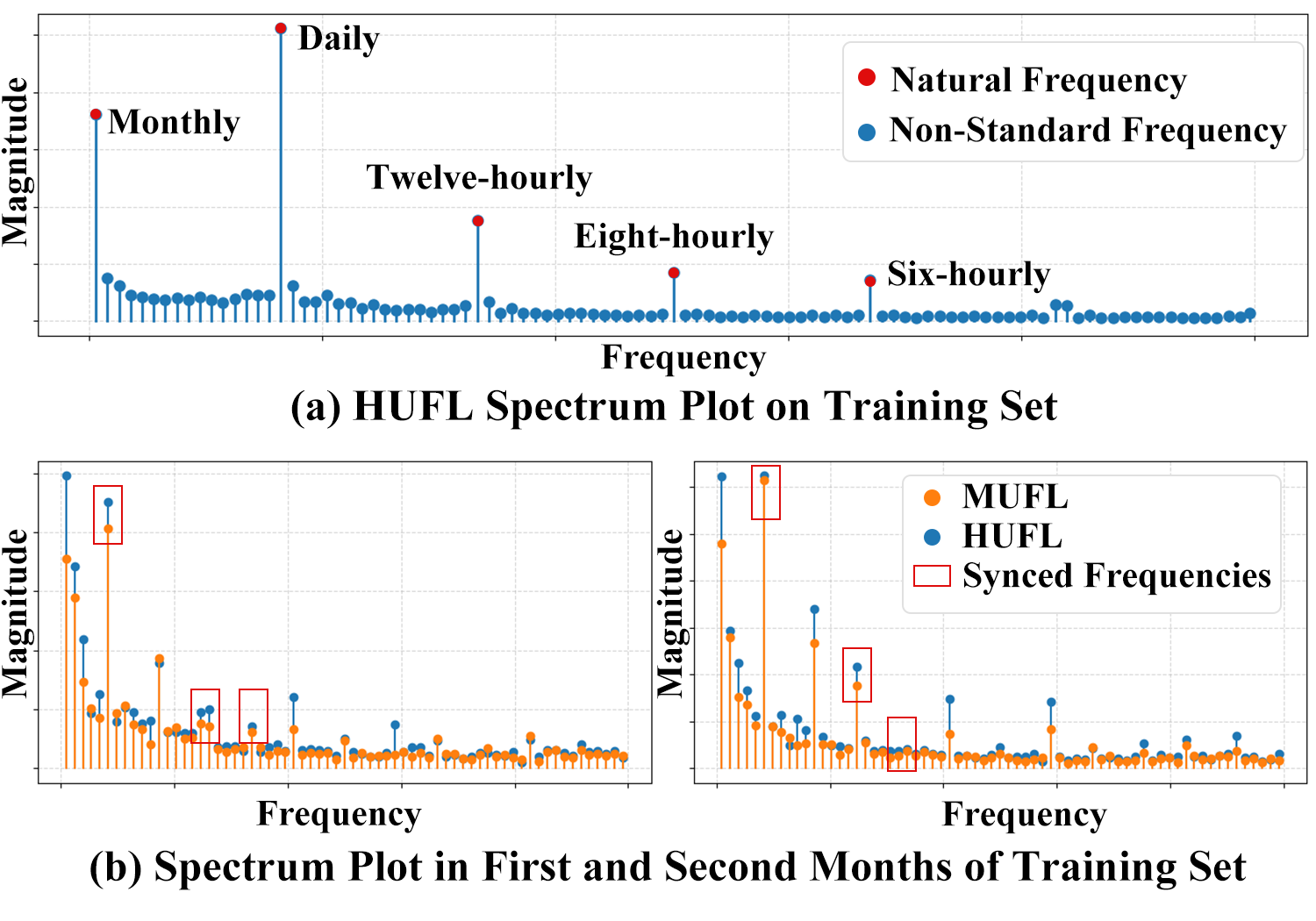}
\caption{Illustration of Stable and Synchronized Variable Frequency Components Using the ETTh1 Dataset. (a) displays the frequency spectrum of the HUFL variable across the entire training set, highlighting dominant stable frequency components with red dots. (b) shows synchronized variable frequency components between HUFL and MUFL variables within a rapidly fluctuating input window (outlined with a red rectangle), after subtracting dominant stable frequency components.}

\label{fig1}
\end{figure}

Traditional time series analysis methods predominantly focus on time-domain analysis \cite{zhou2021informer, nie2023a, liu2024itransformer}, yet they often fall short in capturing periodic information directly\cite{wang2023mppn}. Recent studies have increasingly applied frequency domain analysis to time series, which transforms time series into the frequency space, allowing distinct frequency components to be clearly separated \cite{xu2023fits, yi2024frequency}. Despite these advancements, these methods sometimes fail to effectively differentiate the importance of various frequency components, treating all components equally, which might lead to model overfitting while neglecting crucial a priori periodic information and inter-variable dependencies.

To address these challenges, an effective strategy is the selective extraction and emphasis of frequency components that are most predictive of future observations. We categorize features in time series into (a) stable frequency components and (b) variable frequency components based on their behavior in the frequency domain. Stable components, such as natural cycles (daily, weekly, monthly) and frequencies associated with specific business processes, often appear as dominant frequencies in the series, as determined through frequency domain analysis on training data, depicted in Figure 1a. Variable components, whose characteristics may change over time due to environmental or external factors, are often elusive with static filters. Capturing these dynamically changing frequency components effectively remains a research gap. We hypothesize, based on experimental observations, that these variable frequency components are not only present within individual variables but may also be shared across different variables in a multivariate time series \cite{zhao2024rethinking}. Specifically, there may be synchronization in frequency and intensity changes among these variables, as shown in Figure 1b, revealing underlying inter-variable connections crucial for enhancing multivariate forecasting models.

Based on the foregoing analysis, we introduce FilterTS, an innovative multivariate time series forecasting model that enhances and precisely extracts frequency components through carefully designed filtering modules. The model operates in the frequency domain, converting time-domain convolution to frequency-domain multiplication to improve computational efficiency. FilterTS utilizes two types of filtering modules: 

The Static Global Filtering Module are constructed by performing frequency analysis on the entire training set, building band-pass filters at frequencies corresponding to components with relatively high amplitudes, thereby capturing the dominant stable frequency components. The Dynamic Cross-variable Filtering Module, on the other hand, treats each variable as a filter for the others, dynamically extracting shared frequency components across variables within each lookback window. This approach enhances the capture of variable frequency components. The output sequences from these filters are then merged by a complex matrix.

The experimental results detailed in this paper substantiate the exceptional performance of the FilterTS model across eight real datasets, FilterTS demonstrates superior forecasting accuracy and computational efficiency compared to existing state-of-the-art methods.
The main contributions of this paper are summarized as follows:

\begin{itemize}
    \item We introduce FilterTS, a novel multivariate time series forecasting model leveraging filters to enhance frequency component extraction and improve prediction accuracy.
    \item We develop the Static Global Filtering Module, designed to capture stable periodic components, and the Dynamic Cross-Variable Filtering Module, which dynamically extracts and emphasizes significant frequency components shared across variables.
    \item Our empirical evaluation across eight real-world datasets demonstrates that FilterTS surpasses existing methods in forecasting accuracy and computational efficiency.
\end{itemize}

\section{Related Work}

\subsection{Multivariate Time Series Forecasting Models}
Multivariate Time Series (MTS) involves a set of simultaneously sampled time series data. Commonly, MTS forecasting models attempt to capture dependencies among variables, employing methods such as MLPs \cite{shao2022spatial, ekambaram2023tsmixer, zeng2023transformers, zhang2023crossformer, huang2024hdmixer}, CNNs \cite{wu2022timesnet, donghao2024moderntcn}, GNNs \cite{wu2020connecting, cao2020spectral, cai2024msgnet}, and Transformers \cite{zhou2021informer, liu2024itransformer} to learn these relationships. However, empirical evidence suggests that models that do not explicitly model these dependencies can still achieve strong performance \cite{nie2023a, zhou2023one, xu2023fits}. This may be attributed to the tendency of models that explicitly account for inter-variable dependencies to overfit complex variable interactions when the available prior knowledge is insufficient \cite{nie2023a}.

\subsection{Frequency-Domain Enhanced Time Series Forecasting Models}
Recent developments in time series forecasting have increasingly leveraged frequency-domain analysis to improve predictive accuracy. This approach focuses on extracting and utilizing periodic and global dependencies that are often more discernible in the frequency domain than in the time domain.
FEDformer \cite{zhou2022fedformer} revolutionizes forecasting by applying self-attention directly in the frequency domain, shifting the focus from time-domain dynamics to spectral characteristics. FreTS \cite{yi2024frequency} optimizes predictions by utilizing frequency-domain MLPs to capture static frequency features. FITS \cite{xu2023fits} simplifies models by selectively filtering out high-frequency noise, maintaining only essential low-frequency information.
Additionally, TSLANet \cite{eldele2024tslanet} incorporates adaptive high-frequency filtering techniques to mitigate noise. Despite these technological advancements, a prevalent limitation within these models is their primary focus on static frequency features, which neglects the dynamic changes among variables within the frequency domain.

\section{Preliminary}

\begin{figure*}[t!]
\centering
\includegraphics[width=0.98\textwidth]{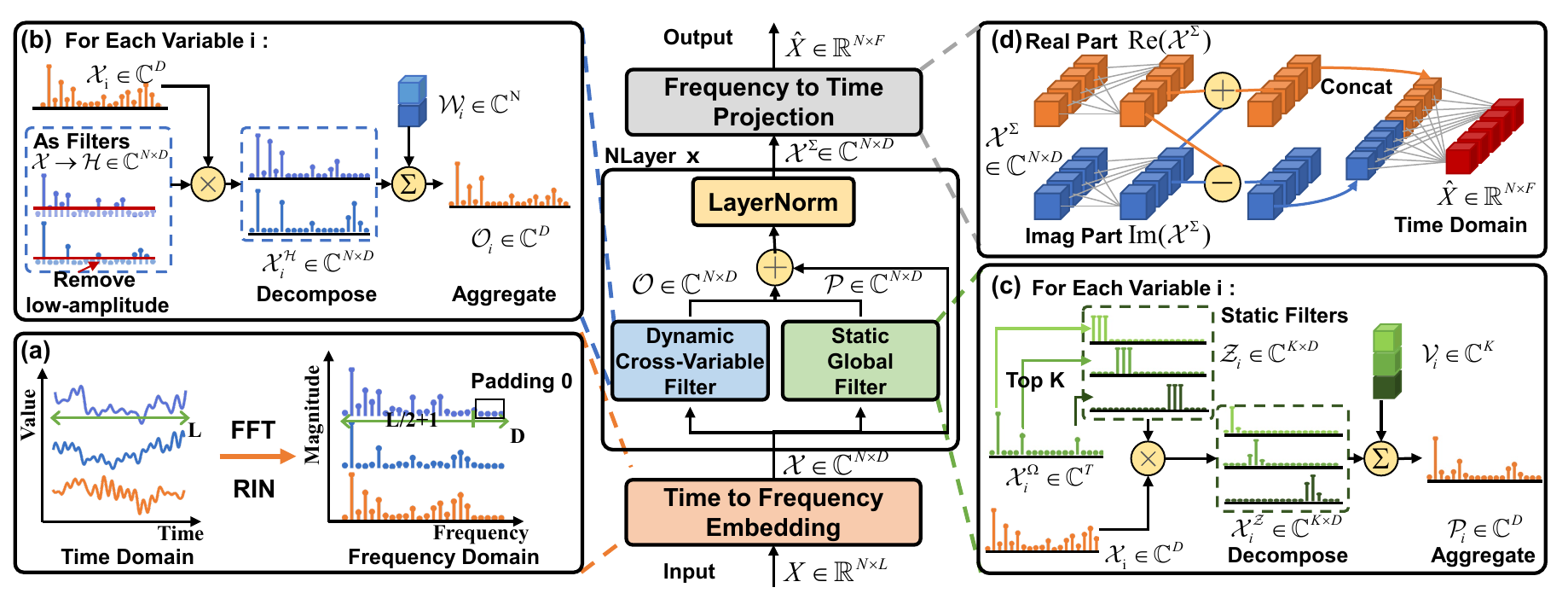} % Reduce the figure size so that it is slightly narrower than the column.
\caption{Overview of FilterTS, which comprises the following key modules: (a) the Time to Frequency Embedding Module, responsible for converting the time-domain series into the frequency domain; (b) the Dynamic Cross-Variable Filtering Module, designed to extract variable frequency components; (c) the Static Global Filtering Module, which focuses on capturing stable frequency components; and finally, (d) the Frequency to Time Projection Module, which transforms the frequency domain representation back into the time domain to generate the final predictions.
}
\label{fig2}
\end{figure*}

\subsection{Problem Formulation}

Given a multivariate time series $X \in \mathbb{R}^{N \times L}$, where $N$ denotes the number of variables and $L$ the lookback window length, the task of multivariate time series forecasting is to predict the values in the future forecasting window. Specifically, for each variable $i \in \{1, \ldots, N\}$, the series is represented as $X_{i} = [x_{i, 1}, \ldots, x_{i, L}]^T$. The objective is to estimate the future values $\hat{X} \in \mathbb{R}^{N \times F}$, where $F$ denotes the length of the forecasting window. The forecast values for each variable are given by $\hat{X}_i = [\hat{x}_{i, L+1}, \ldots, \hat{x}_{i, L+F}]^T$ for $i \in \{1, \ldots, N\}$.

\subsection{Fast Fourier Transform (FFT)}

The Fast Fourier Transform (FFT)\cite{duhamel1990fast} is an efficient algorithm for computing the Discrete Fourier Transform (DFT) of a sequence. Consider a univariate time series \( z_t \in \mathbb{R}^L \), where \( t \) represents the time index, and \( L \) is the length of the sequence. The FFT transforms this time series from the time domain into its frequency domain representation \( \mathcal{Z}_f \in \mathbb{C}^L \), where \( f \) represents the frequency index. The transformation is expressed as:

\begin{equation}
\mathcal{Z}_f = \sum_{t=0}^{L-1} z_t \cdot e^{-j \frac{2\pi}{L} f t}, \quad f = 0, 1, \dots, L-1
\end{equation}

where \( j \) is the imaginary unit.
The frequency domain representation \( \mathcal{Z}_f \) can be decomposed into its real and imaginary components, denoted as \( \text{Re}(\mathcal{Z}_f) \) and \( \text{Im}(\mathcal{Z}_f) \), respectively.
The magnitude (or amplitude) of \( \mathcal{Z}_f \), representing the strength of the frequency components, is defined as:

\begin{equation}
|\mathcal{Z}_f| = \sqrt{\text{Re}(\mathcal{Z}_f)^2 + \text{Im}(\mathcal{Z}_f)^2}
\end{equation}

\section{Method}
\subsection{Overall Structure}

Figure \ref{fig2} illustrates the comprehensive architecture of the FilterTS model, comprised of four main components: the Time to Frequency Embedding Module (\text{T\textit{2}FEmbed}), the Dynamic Cross-Variable Filtering Module (DCFilter), the Static Global Filtering Module (SGFilter), and the Frequency to Time Projection Layer (\text{F\textit{2}TOut}).

We first transform the given multivariate time series \( X \in \mathbb{R}^{N \times L} \) into a frequency domain representation \( \mathcal{X} \) to facilitate subsequent filtering operations:
\begin{equation}
\mathcal{X} = \text{T\textit{2}FEmbed}(X)
\end{equation}
The resulting \( \mathcal{X} \in \mathbb{C}^{N \times D} \) represents the data in a frequency domain format optimized for neural network processing, where \( \mathbb{C} \) denotes the complex domain and \( D \) represents the dimension of hidden layers.

The frequency domain representation \( \mathcal{X} \) is then concurrently fed into two distinct pathways: 
\begin{align}
\mathcal{O} &= \text{DCFilter}(\mathcal{X}) \\
\mathcal{P} &= \text{SGFilter}(\mathcal{X})
\end{align}
where \(\text{DCFilter}(\cdot)\) and \(\text{SGFilter}(\cdot)\) denote the Dynamic
Cross-Variable Filtering Module and Static Global Filtering Module, respectively. 
Both \( \mathcal{O} \) and \( \mathcal{P} \) are frequency domain representations in \( \mathbb{C}^{N \times D} \), corresponding to the cross-variable and stable filtering outputs.

Subsequently, the outputs from the embedding and both filtering modules are aggregated:
\begin{equation}
\mathcal{X}^{\Sigma} = \text{LayerNorm}(\alpha \mathcal{X} + \beta \mathcal{O} + \gamma \mathcal{P})
\end{equation}
where \( \alpha \), \( \beta \), and \( \gamma \) are weighting coefficients optimized during training. We apply LayerNorm to stabilize the training process.

This process can be repeated for \( e \) layers, where equations (4) to (6) are repeated \( e \) times. In each iteration, the output \( \mathcal{X}^{\Sigma} \) from the previous layer serves as the input \( \mathcal{X} \) in equation (4) for the next layer.

Finally, the aggregated frequency domain representation \( \mathcal{X}^{\Sigma} \) is fed into the Frequency-to-Time Projection Layer\(\text{F\textit{2}TOut}(\cdot)\), which converts the frequency domain back into the time domain, generating forecasts \( \hat{X} \in \mathbb{R}^{N \times F} \) for \( F \) future steps:
\begin{equation}
\hat{X} = \text{\text{F\textit{2}TOut}}(\mathcal{X}^{\Sigma})
\end{equation}

\subsection{Time to Frequency Embedding}

The Time-to-Frequency Embedding Module converts the time-domain input series into the frequency domain, leveraging the equivalence between time-domain convolution and frequency-domain multiplication to enhance computational efficiency in subsequent filtering stages.

Firstly, we apply instance normalization to each time series \( X_{i} \in \mathbb{R}^L \) to mitigate the effects of varying data distributions \cite{kim2022reversible}:
\begin{equation}
\tilde{X}_i = \frac{X_i - \mu(X_i)}{\sigma(X_i) + \epsilon}
\end{equation}
where $\mu(X_i)$ and $\sigma(X_i)$ are the mean and standard deviation of each individual time series, and $\epsilon$ is a small constant to ensure numerical stability.

Secondly, the Fast Fourier Transform (FFT) is applied to convert the normalized time series into the frequency domain efficiently:
\begin{equation}
\mathcal{X}_i^{+} = \text{FFT}\left(\tilde{X}_i \, \| \, \text{zeros}{(L)}\right)
\end{equation}
In this formulation, \(\tilde{X}_i\) is zero-padded by appending \(L\) zeros, increasing the sequence length to \(2L\). This ensures that the frequency domain multiplication corresponds to linear convolution in the time domain, rather than circular convolution. The FFT then transforms this extended sequence into a complex number spectrum \(\mathcal{X}_i^{+} \in \mathbb{C}^{2L}\). For an in-depth introduction to FFT and extended-length effects, please see the Appendix.

Finally, to match the required model dimensionality \( D \), while maintaining fidelity to the frequency domain representation, Fourier interpolation is performed:
\begin{equation}
\mathcal{X}_{i,j} = \left\{ \begin{array}{cl}
\mathcal{X}_{i,j}^{+} & \text{if } j < L+1 \\
0 & \text{if } j \geq L+1
\end{array} \right. \quad \text{for } j = 1, \dots, D
\end{equation}

Given the conjugate symmetry of the FFT for real-valued signals, only the first \(L+1\) components contain unique information. Therefore, this operation either extends the truncated frequency spectrum by zero-padding if \( D \) exceeds \( L+1 \) or reduces it by selecting only the first \( D \) low-frequency components.

\subsection{Dynamic Cross-Variable Filtering Module}

\subsubsection{Time Series as Filter}

The foundational theory of utilizing time series as finite impulse response (FIR) filters is grounded in the signal processing principle that the filtering operation can be represented as convolution in the time domain or multiplication in the frequency domain. This section discusses the transformation of a time series into an FIR filter and its practical implications.

An FIR filter is characterized by a finite sequence of \( N \) coefficients \( h[m] \), defining the filter's impulse response in the time domain:
\begin{equation}
y[n] = \sum_{m=0}^{M-1} h[m] \cdot x[n-m]
\end{equation}
The output signal \( y[n] \) is determined by convolving the input signal \( x[n] \) with the filter coefficients \( h[m] \), where \( M \) indicates the memory of the filter. 

Consider two time series, each of length \( L \): one representing the input signal \( x = [x_{1}, \dots, x_{L}] \) and the other acting as the filter \( h = [h_{1}, \dots, h_{L}] \). The filtered output signal \( y \) of the FIR filter is obtained by convolving these two sequences:
\begin{equation}
y_{n} = \sum_{m=1}^{L} h_{m} \cdot x_{n-m}, \quad \text{for } n = 1, \dots, L
\end{equation}
where \( y_{n} \) denotes the filtered time series at time \( n \).

The convolution theorem facilitates this operation by transforming the time-domain convolution into a frequency-domain multiplication:
\begin{equation}
Y_{f} = X_{f} \cdot H_{f}, \quad \text{for } f = 1, \dots, L
\end{equation}
where \( Y_{f}, X_{f}, \) and \( H_{f} \) are the discrete Fourier transforms of \( \mathbf{y}, \mathbf{x} \) and \( \mathbf{h} \) at frequency \(f\), respectively.

This principle enables significant frequency components in \( X_{f} \) and \( H_{f} \) to produce pronounced energy in  \( Y_{f} \), thereby accentuating the frequencies common to both sequences. This characteristic is particularly advantageous for the analysis of interactions within the frequency domain.

\subsubsection{Generation of Dynamic Cross-Variable Filters}

Building on the concept of utilizing time series as filters, we leverage the frequency domain representation \( \mathcal{X} \in \mathbb{C}^{N \times D} \) to generate dynamic filters. In this context, dynamic refers to the construction of filters that adapt within each look-back window, capturing key frequency components that are shared across variables. 

Dynamic filters are derived by evaluating the amplitude of each frequency component. To focus on significant frequencies and reduce the influence of noise, components below a certain amplitude threshold are discarded. This threshold is dynamically set based on the amplitude distribution of each variable's frequency components:
\begin{equation}
    \tau_{i} = \text{quantile}(\left| \mathcal{X}_i \right|, \alpha)
\end{equation}
where \( \tau_{i} \) represents the \( \alpha \)-quantile, defining the minimum amplitude threshold for the \( i \)-th variable to consider a frequency component significant.

The dynamic filter for each frequency component \( \mathcal{H}_{i, f} \) is then defined as:
\begin{equation}
    \mathcal{H}_{i, f} = \begin{cases} 
    \mathcal{X}_{i, f} & \text{if } \left| \mathcal{X}_{i, f} \right| > \tau_{i} \\
    0 & \text{otherwise}
    \end{cases}
\end{equation}

where \( \left| \mathcal{X}_{i, f} \right| \) representing the magnitude of the \( f \)-th frequency component for the \( i \)-th variable.
The resulting dynamic filters, which capture key frequency components of each variable's real-time input, are combined to filter each variable's original series to enhance shared frequency components across variables. These dynamic filters are assembled into a matrix \( \mathcal{H} = [\mathcal{H}_1, \ldots, \mathcal{H}_N] \) of dimensions \( \mathbb{C}^{N \times D} \).
\subsubsection{Filtering Decomposition and Aggregation}

After obtaining the dynamic cross-variable filters \( \mathcal{H} \), we return to the processing of the input time series. Upon adjusting the amplitude and phase of the frequency domain representations using the scaling matrix \( \mathcal{A}^{o} \in \mathbb{C}^{N \times D} \), we obtain the adjusted frequency domain representation \( \mathcal{X}^{o} \):
\begin{equation}
    \mathcal{X}^{o} = \mathcal{X} \odot \mathcal{A}^{o}
\end{equation}

For each variable \( i \) and each dynamic filter \( k \), the adjusted representation is multiplied by the respective filter to obtain:
\begin{equation}
    \mathcal{X}_{i, k}^{\mathcal{H}} = \mathcal{X}^{o}_i \odot H^{*}_{k}
\end{equation}
where \( \odot \) denotes the element-wise multiplication. \( H^{*}_{k} \) is the conjugate of the \( k \)-th dynamic filter. Here, \( \mathcal{X}_{i,k}^{\mathcal{H}} \in \mathbb{C}^{D} \) represents the subsequence generated by filtering the time series \( i \) using the dynamic filter derived from time series \( k \).

Next, aggregation of these filtered subsequences is performed using a complex weight matrix \( \mathcal{W} \in \mathbb{C}^{N \times N} \), which is processed through modified softmax and ReLU functions to form a sparse matrix:
\begin{equation}
    \mathcal{W}^* = \text{softmax}(\text{ReLU}(\mathcal{W}))
\end{equation}
The weighted aggregation for each variable \( i \) is then computed as follows:
\begin{equation}
    \mathcal{O}_i = \sum_{k=1}^N \mathcal{X}_{i, k}^{\mathcal{H}} \cdot {W}^{*}_{i, k}
\end{equation}
\begin{equation}
    \mathcal{O} = [\mathcal{O}_{1}, \ldots, \mathcal{O}_{N}]
\end{equation}

Here, \( \mathcal{O} \in \mathbb{C}^{N \times D} \) represents the output of the Dynamic Cross-Variable Filtering Module. It enhances shared frequency components across variables, allowing the model to effectively capture and utilize critical inter-variable interactions through dynamic filtering.

\subsection{Static Global Filtering Module}
   
The Static Global Filtering Module identifies dominant stable frequency components across the entire training dataset. By constructing band-pass filters targeting the top \( K \) high-amplitude frequencies, it extracts key frequency representations from each input sequence during filtering.

\subsubsection{Generation of Static Global Filters}

Let the training dataset \( X^{\Omega} \in \mathbb{R}^{N \times T} \) consist of multivariate time series, where \( N \) is the number of variables and \( T \) is the length of each sequence.
The Fourier transform is applied  to obtain the global frequency domain representation \( \mathcal{X}^{\Omega} \in \mathbb{C}^{N \times T} \):

\begin{equation}
    \mathcal{X}^{\Omega} = \text{FFT}(X^{\Omega})
\end{equation}

Since the model's input sequence \( X \in \mathbb{R}^{N \times L} \) has a lookback window length \( L \), the frequency resolution of \( \mathcal{X}^{\Omega} \) (of length \( T \)) is higher than that of the input sequence. To align the frequency resolutions, we perform down-sampling on \( \mathcal{X}^{\Omega} \) to reduce its length from \( T \) to \( L \). For each variable \( i \), the down-sampled frequency representation is obtained by summing the magnitudes of frequency components over non-overlapping windows \( \kappa
 \) of size \( {T}/{L} \):

\begin{equation}
    \tilde{\mathcal{X}}^{\Omega}_{i, f} = \sum_{m=\kappa \times f}^{\kappa
\times(f+1) -1} \mathcal{X}^{\Omega}_{i, m}
\end{equation}
where \( \tilde{\mathcal{X}}^{\Omega}_{i,f} \) represents the downsampled magnitude of frequency component \( f \) for variable \( i \), with \( i \in \{1, \dots, N\} \) and \( f \in \{1, \dots, L\} \).

For each variable \( \tilde{\mathcal{X}}^{\Omega}_{i} \), the top \( K \) frequency components are selected based on their magnitudes. Let \( \{f^{*}_{i, 1}, f^{*}_{i, 2}, \dots, f^{*}_{i, K}\} \) denote the indices of these top \( K \) frequencies for variable \( i \). For each selected frequency \( f^{*}_{i, s} \), define a band-pass filter \( \mathcal{Z}_{i, s} \) as:

\begin{equation}
    \mathcal{Z}_{i, s, f^{\prime}} = 
    \begin{cases}
        1 & \text{if } f^{\prime} \in [f^{*}_{i, s} - \Delta f, f^{*}_{i, s} + \Delta f] \\
        0 & \text{otherwise}
    \end{cases}
\end{equation}

where \( \Delta f \) represents the half bandwidth of the filter, accounting for minor deviations around the dominant frequency, and \(f^{\prime} \in \{ 1, \ldots, D \} \) to match the dimensionality of the input frequency domain representation \( \mathcal{X}_i \in \mathbb{C}^{D} \), as both the filter \( \mathcal{Z}_{i, s} \) have the same dimensions \( \mathbb{C}^{D} \).

These static filters of variable \(i\) are assembled into a matrix \(\mathcal{Z}_{i} = [\mathcal{Z}_{i, 1}, \ldots, \mathcal{Z}_{i, K}] \) of dimensions \( \mathbb{C}^{K \times D} \).

\subsubsection{Filtering Decomposition and Aggregation}
The frequency domain representation of the model's multivariate input \( \mathcal{X} \in \mathbb{C}^{N \times D} \), obtained through Fourier transform, is first adjusted in amplitude and phase using an amplitude scaling matrix \( \mathcal{A}^{p} \in \mathbb{C}^{N \times D} \):

\begin{equation}
    \mathcal{X}^{p} = \mathcal{X} \odot \mathcal{A}^{pend}
\end{equation}

For each variable \( i \) and each static filter \( s \) in \( \mathcal{Z}_{i} \), the adjusted frequency representation is filtered by multiplying it with the corresponding static filter:

\begin{equation}
    \mathcal{X}_{i, k}^{\mathcal{Z}} = \mathcal{X}^{p}_{i} \odot \mathcal{Z}_{i,s}
\end{equation}

where \( \odot \) denotes element-wise multiplication, and \( \mathcal{Z}_{i,s} \) is the static filter \( s \) for variable \( i \).

To aggregate the filtered representations, a complex weight matrix \( \mathcal{V} \in \mathbb{C}^{N \times K} \) is applied, which is first processed through a modified softmax and ReLU function to enforce sparsity:

\begin{equation}
    \mathcal{V}^* = \text{softmax}(\text{ReLU}(\mathcal{V}))
\end{equation}

The weighted aggregation for each variable is then computed through matrix multiplication:

\begin{equation}
    \mathcal{P}_{i} = \sum_{s=1}^K \mathcal{X}_{i, k}^{\mathcal{Z}} \cdot V^{*}_{i,s}
\end{equation}

Finally, the static filtered sequences for all variables are assembled into the output matrix \( \mathcal{P} = [\mathcal{P}_{1}, \dots, \mathcal{P}_{N}] \) as the static global filtering module's output.
This sequence of operations enhances static frequency components within the data, effectively leveraging the inherent static frequencies of the sequences for robust feature extraction.
\begin{table*}[t!]
\centering
\resizebox{0.9\textwidth}{!}{\begin{tabular}{cc|cc|cc|cc|cc|cc|cc|cc}
\toprule
 \multicolumn{2}{c}{Models}  & \multicolumn{2}{c}{Ours} & \multicolumn{2}{c}{TimeMixer} & \multicolumn{2}{c}{iTransformer} & \multicolumn{2}{c}{PatchTST} & \multicolumn{2}{c}{Crossformer} & \multicolumn{2}{c}{TimesNet} & \multicolumn{2}{c}{FreTS} \\
\cmidrule(lr){1-2} \cmidrule(lr){3-4} \cmidrule(lr){5-6} \cmidrule(lr){7-8} \cmidrule(lr){9-10} \cmidrule(lr){11-12} \cmidrule(lr){13-14} \cmidrule(lr){15-16} 
\multicolumn{2}{c}{Metric} & MSE & MAE & MSE & MAE & MSE & MAE & MSE & MAE & MSE & MAE & MSE & MAE & MSE & MAE \\
\midrule
\multirow{5}{*}{\rotatebox[origin=c]{90}{ETTm1}} 
& 96  & \textbf{0.321} & \textbf{0.360} & 0.333 & 0.371 & 0.334 & 0.368 & \underline{0.329} & \underline{0.367} & 0.404 & 0.426 & 0.338 & 0.375 & 0.335 & 0.371 \\
& 192 & \textbf{0.363} & \textbf{0.382} & 0.367 & 0.386 & 0.377 & 0.391 & \underline{0.367} & \underline{0.385} & 0.450 & 0.451 & 0.374 & 0.387 & 0.377 & 0.394 \\
& 336 & \textbf{0.395} & \textbf{0.403} & \underline{0.397} & \underline{0.408} & 0.426 & 0.420 & 0.399 & 0.410 & 0.532 & 0.515 & 0.410 & 0.411 & 0.413 & 0.418 \\
& 720 & 0.462 & \textbf{0.438} & \underline{0.460} & 0.445 & 0.491 & 0.459 & \textbf{0.454} & \underline{0.439} & 0.666 & 0.589 & 0.478 & 0.450 & 0.483 & 0.461 \\
& Avg & \textbf{0.385} & \textbf{0.396} & 0.389 & 0.403 & 0.407 & 0.410 & \underline{0.387} & \underline{0.400} & 0.513 & 0.495 & 0.400 & 0.406 & 0.402 & 0.411 \\

\midrule
\multirow{5}{*}{\rotatebox[origin=c]{90}{ETTm2}} 
& 96  & \textbf{0.172} & \textbf{0.255} & \underline{0.174} & \underline{0.258} & 0.180 & 0.264 & 0.175 & 0.259 & 0.287 & 0.366 & 0.187 & 0.267 & 0.181 & 0.269 \\
& 192 & \textbf{0.237} & \textbf{0.299} & \underline{0.239} & \underline{0.302} & 0.250 & 0.309 & 0.241 & \underline{0.302} & 0.414 & 0.492 & 0.249 & 0.309 & 0.249 & 0.322 \\
& 336 & \underline{0.299} & \textbf{0.398} & \textbf{0.296} & \underline{0.340} & 0.311 & 0.348 & 0.305 & 0.343 & 0.597 & 0.542 & 0.321 & 0.351 & 0.340 & 0.382 \\
& 720 & \underline{0.397} & \textbf{0.394} & \textbf{0.393} & \underline{0.397} & 0.412 & 0.407 & 0.402 & 0.400 & 1.730 & 1.042 & 0.408 & 0.403 & 0.449 & 0.455 \\
& Avg & \textbf{0.276} & \textbf{0.321} & \textbf{0.276} & \underline{0.324} & 0.288 & 0.332 & 0.281 & 0.326 & 0.757 & 0.611 & 0.291 & 0.333 & 0.305 & 0.357 \\

\midrule
\multirow{5}{*}{\rotatebox[origin=c]{90}{ETTh1}}
                    & 96  & \textbf{0.374} & \textbf{0.391} & \underline{0.384} & \underline{0.398} & 0.386 & 0.405 & 0.414 & 0.419 & 0.423 & 0.448 & \underline{0.384} & 0.402 & 0.390 & 0.404 \\
                    & 192 & \textbf{0.424} & \textbf{0.421} & 0.439 & \underline{0.429} & 0.441 & 0.436 & 0.460 & 0.445 & 0.471 & 0.474 & \underline{0.436} & \underline{0.429} & 0.448 & 0.439 \\
                    & 336 & \textbf{0.464} & \textbf{0.441} & \underline{0.487} & \underline{0.455} & \underline{0.487} & 0.458 & 0.501 & 0.466 & 0.570 & 0.546 & 0.491 & 0.469 & 0.501 & 0.470 \\
                    & 720 & \textbf{0.470} & \textbf{0.466} & 0.502 & \underline{0.482} & 0.503 & 0.491 & \underline{0.500} & 0.488 & 0.653 & 0.621 & 0.521 & 0.500 & 0.559 & 0.535 \\
                    & Avg & \textbf{0.433} & \textbf{0.430} & \underline{0.453} & \underline{0.441} & 0.454 & 0.448 & 0.469 & 0.455 & 0.529 & 0.522 & 0.458 & 0.450 & 0.475 & 0.462 \\
\midrule
\multirow{5}{*}{\rotatebox[origin=c]{90}{ETTh2}}
                    & 96  & \textbf{0.290} & \textbf{0.338} & \underline{0.293} & \underline{0.344} & 0.297 & 0.349 & 0.302 & 0.348 & 0.745 & 0.584 & 0.340 & 0.374 & 0.317 & 0.373 \\
                    & 192 & \textbf{0.374} & \textbf{0.390} & \textbf{0.374} & \underline{0.396} & 0.380 & 0.400 & 0.388 & 0.400 & 0.877 & 0.656 & 0.402 & 0.414 & 0.427 & 0.442 \\
                    & 336 & \textbf{0.406} & \textbf{0.420} & \underline{0.416} & \underline{0.430} & 0.428 & 0.432 & 0.426 & 0.433 & 1.043 & 0.731 & 0.452 & 0.452 & 0.526 & 0.505 \\
                    & 720 & \textbf{0.418} & \textbf{0.437} & 0.440 & 0.452 & \underline{0.427} & \underline{0.445} & 0.431 & 0.446 & 1.104 & 0.763 & 0.462 & 0.468 & 0.684 & 0.591 \\
                    & Avg & \textbf{0.372} & \textbf{0.396} & \underline{0.381} & \underline{0.406} & 0.383 & 0.407 & 0.387 & 0.407 & 0.942 & 0.684 & 0.414 & 0.427 & 0.489 & 0.478 \\
\midrule
\multirow{5}{*}{\rotatebox[origin=c]{90}{Electricity}} 
                    & 96  & \underline{0.151} & \underline{0.245} & 0.157 & 0.250 & \textbf{0.148} & \textbf{0.240} & 0.181 & 0.270 & 0.219 & 0.314 & 0.168 & 0.272 & 0.177 & 0.264 \\
                    & 192 & \underline{0.163} & \underline{0.256} & 0.177 & 0.263 & \textbf{0.162} & \textbf{0.253} & 0.188 & 0.274 & 0.231 & 0.322 & 0.184 & 0.289 & 0.183 & 0.271 \\
                    & 336 & \underline{0.180} & \underline{0.274} & 0.189 & 0.282 & \textbf{0.178} & \textbf{0.269} & 0.204 & 0.293 & 0.246 & 0.337 & 0.198 & 0.300 & 0.198 & 0.288 \\
                    & 720 & \underline{0.224} & \textbf{0.311} & 0.233 & \underline{0.316} & 0.225 & 0.317 & 0.246 & 0.324 & 0.280 & 0.363 & \textbf{0.220} & 0.320 & 0.234 & 0.322 \\
                    & Avg & \underline{0.180} & \underline{0.271} & 0.189 & 0.278 & \textbf{0.178} & \textbf{0.270} & 0.205 & 0.290 & 0.244 & 0.334 & 0.193 & 0.295 & 0.255 & 0.300 \\
\midrule
\multirow{5}{*}{\rotatebox[origin=c]{90}{Exchange}} 
                    & 96  & \textbf{0.081} & \textbf{0.199} & 0.087 & 0.205 & 0.086 & 0.206 & 0.088 & \underline{0.205} & 0.256 & 0.367 & 0.107 & 0.234 & \underline{0.085} & 0.211 \\
                    & 192 & \textbf{0.171} & \textbf{0.294} & 0.179 & 0.300 & 0.177 & \underline{0.299} & \underline{0.176} & \underline{0.299} & 0.470 & 0.509 & 0.226 & 0.344 & 0.181 & 0.312 \\
                    & 336 & \underline{0.321} & \underline{0.409} & 0.333 & 0.417 & 0.331 & 0.417 & \textbf{0.301} & \textbf{0.397} & 1.268 & 0.883 & 0.367 & 0.448 & 0.471 & 0.508 \\
                    & 720 & \textbf{0.837} & \textbf{0.688} & 0.912 & 0.719 & \underline{0.847} & \underline{0.691} & 0.901 & 0.714 & 1.767 & 1.068 & 0.964 & 0.746 & 0.858 & 0.695 \\
                    & Avg & \textbf{0.352} & \textbf{0.397} & 0.378 & 0.410 & \underline{0.360} & \underline{0.403} & 0.367 & 0.404 & 0.940 & 0.707 & 0.416 & 0.443 & 0.399 & 0.432 \\
\midrule
\multirow{5}{*}{\rotatebox[origin=c]{90}{Traffic}} 
                    & 96  & \underline{0.448} & 0.309 & 0.485 & 0.323 & \textbf{0.395} & \textbf{0.268} & 0.462 & 0.295 & 0.522 & \underline{0.290} & 0.593 & 0.321 & 0.512 & 0.328 \\
                    & 192 & \underline{0.455} & 0.307 & 0.488 & 0.322 & \textbf{0.417} & \textbf{0.276} & 0.466 & 0.296 & 0.530 & \underline{0.293} & 0.617 & 0.336 & 0.507 & 0.325 \\
                    & 336 & \underline{0.472} & 0.313 & 0.507 & 0.321 & \textbf{0.433} & \textbf{0.283} & 0.482 & \underline{0.304} & 0.558 & 0.305 & 0.629 & 0.336 & 0.521 & 0.331 \\
                    & 720 & \underline{0.508} & 0.332 & 0.549 & 0.335 & \textbf{0.467} & \textbf{0.302} & 0.514 & \underline{0.322} & 0.589 & 0.328 & 0.640 & 0.350 & 0.562 & 0.348 \\
                    & Avg & \underline{0.471} & 0.315 & 0.507 & 0.325 & \textbf{0.428} & \textbf{0.282} & 0.481 & \underline{0.304} & 0.550 & \underline{0.304} & 0.620 & 0.336 & 0.526 & 0.333 \\
\midrule
\multirow{5}{*}{\rotatebox[origin=c]{90}{Weather}} 
                    & 96  & \underline{0.162} & \textbf{0.207} & 0.166 & \underline{0.213} & 0.174 & 0.214 & 0.177 & 0.218 & \textbf{0.158} & 0.230 & 0.172 & 0.220 & 0.182 & 0.236 \\
                    & 192 & \underline{0.209} & \underline{0.252} & \underline{0.209} & \textbf{0.251} & 0.221 & 0.254 & 0.225 & 0.259 & \textbf{0.206} & 0.277 & 0.219 & 0.261 & 0.219 & 0.217 \\
                    & 336 & \textbf{0.263} & \textbf{0.292} & \underline{0.264} & \underline{0.293} & 0.278 & 0.296 & 0.278 & 0.297 & 0.272 & 0.335 & 0.280 & 0.306 & 0.270 & 0.313 \\
                    & 720 & \underline{0.344} & \underline{0.344} & \textbf{0.342} & \textbf{0.343} & 0.358 & 0.349 & 0.354 & 0.348 & 0.398 & 0.418 & 0.365 & 0.359 & 0.348 & 0.380 \\
                    & Avg & \textbf{0.244} & \textbf{0.274} & \underline{0.245} & \underline{0.275} & 0.258 & 0.278 & 0.259 & 0.281 & 0.259 & 0.315 & 0.259 & 0.287 & 0.255 & 0.300 \\
\midrule
\multicolumn{2}{c}{1st Count} & \textbf{23} & \textbf{28} & 5 & 2 & \underline{9} & \underline{9} & 2 & 1 & 2 & 0 & 1 & 0 & 0 & 0 \\
\bottomrule
\end{tabular}}
\caption{Results of the multivariate long-term time series forecasting task, evaluated using MSE and MAE (lower is better). The lookback window for all models was set to 96. The best results are highlighted in \textbf{bold}, while the second-best results are \underline{underlined}.}
\label{Tab1}
\end{table*}
\subsection{Frequency to Time Projection}

Finally to transform the frequency domain representation \( \mathcal{X}^{\Sigma } \) back into the time domain, we first apply a linear transformation to handle the real and imaginary components of the frequency domain data:
\begin{equation}
\text{Re}(\mathcal{\hat{X}}) = \text{Re}(\mathcal{X}^{\Sigma }) \cdot \mathbf{U}^{\text{Re}} - \text{Im}(\mathcal{X}^{\Sigma }) \cdot \mathbf{U}^{\text{Im}}
\end{equation}
\begin{equation}
\text{Im}(\mathcal{\hat{X}}) = \text{Re}(\mathcal{X}^{\Sigma }) \cdot \mathbf{U}^{\text{Im}} + \text{Im}(\mathcal{X}^{\Sigma }) \cdot \mathbf{U}^{\text{Re}}
\end{equation}
Here, \( \mathbf{U}^{\text{Re}} \) and \( \mathbf{U}^{\text{Im}} \) represent the weight matrices for the real and imaginary components, respectively, in the space \( \mathbb{R}^{D \times D} \). This dual transformation preserves the properties of the complex numbers, such as phase relationships and amplitude variations. Subsequently, the real and imaginary parts are concatenated and then linearly transformed to map the combined complex data into the predicted time series output:
\begin{equation}
X^{\|} = [\text{Re}(\mathcal{\hat{X}}) \, \| \, \text{Im}(\mathcal{\hat{X}})]
\end{equation}
\begin{equation}
\hat{X} = X^{\|} \cdot \mathbf{Q}
\end{equation}
where \( \mathbf{Q} \) is the final linear transformation matrix with dimensions \( \mathbb{R}^{2D \times F} \), and \( \hat{X} \) is the final predicted time series in \( \mathbb{R}^{D \times F} \).

\section{Experiments}
\subsection{Experimental Details}
\subsubsection{Datasets}
Following \cite{zhou2021informer} and \cite{nie2023a}, we evaluate our proposed model on eight widely-recognized multivariate time series forecasting datasets, spanning diverse domains such as energy, economics, transportation, and climate. Specifically, these datasets include the four ETT datasets (ETTh1, ETTh2, ETTm1, ETTm2), Electricity, Exchange Rate, Traffic, and Weather.

\subsubsection{Baselines}
We select several state-of-the-art multivariate time series forecasting models as baselines, including TimeMixer \cite{wang2024timemixer}, iTransformer \cite{liu2024itransformer}, PatchTST \cite{nie2023a}, Crossformer \cite{zhang2023crossformer}, and TimesNet \cite{wu2022timesnet}.  Additionally, we incorporate FreTS \cite{yi2024frequency}, a novel model built upon frequency-domain. 

\subsubsection{Setup}
All experiments were conducted on an NVIDIA GeForce RTX 4090 24GB GPU. We adopted a consistent experimental setup identical to that of iTransformer \cite{liu2024itransformer} to ensure a fair comparison. Specifically, the lookback length for all models was fixed at 96, with prediction lengths set to $F \in \{96, 192, 336, 720\}$, Mean Squared Error (MSE) was used as the loss function. 

For detailed information on datasets, model descriptions, hyperparameter settings, and other experimental implementation specifics, please refer to the Appendix.

\subsection{Main Results}

Table \ref{Tab1} presents the predictive performance of FilterTS across eight multivariate long-term time series forecasting datasets, demonstrating superior accuracy over current state-of-the-art models in most cases. Specifically, averaging MSE across all prediction lengths, FilterTS achieved the best performance on 6 out of the 8 datasets and secured the second-best performance on the remaining 2 datasets. 

The slightly weaker performance of FilterTS on the Traffic and ECL datasets can be attributed to the high number of variables in these datasets, which presents challenges for the static weight matrix in capturing complex inter-variable relationships, particularly under nonlinear conditions. In contrast, models like iTransformer, which utilize a sophisticated attention mechanism, dynamically adjust weights to better handle complex, multivariate interactions.

Despite these challenges, FilterTS reduced the average MSE by 4.24\% compared to PatchTST, a representative of channel-independent models, indicating that effectively leveraging inter-variable information can significantly enhance model performance. Additionally, FilterTS outperformed models that capture inter-variable relationships using MLP (TimeMixer) and attention mechanisms (iTransformer), reducing average MSE by 3.69\% and 2.02\%, respectively. This suggests that the method of extracting and integrating shared frequency components across variables through filtering mechanisms is more effective than simpler fusion strategies.
Furthermore, FilterTS consistently led FreTS across all tasks, underscoring that its performance benefits derive not solely from its construction in the frequency domain, but from its effective filtering strategies.

\subsection{Model Analysis}

\subsubsection{Ablation Study}
To validate the effectiveness of the FilterTS module design, we conducted ablation studies on three datasets. Specifically, we evaluated the following variants: 
\textbf{w/o-SGF}: The Static Global Filtering Module was removed.
\textbf{w/o-DCF}: The Dynamic Cross-Variable Filtering Module was removed.
\textbf{w/o-SGF\&DCF}: Both modules removed.
\textbf{re-MLP}: The Dynamic Cross-Variable Filtering Module was replaced with a simple MLP to capture inter-variable relationships.
\textbf{re-Attn}: The Dynamic Cross-Variable Filtering Module was replaced with an attention mechanism to capture inter-variable relationships.
%\begin{itemize}
%\item \textbf{w/o-SGF}: The Static Global Filtering Module was removed.
%\item \textbf{w/o-DCF}: The Dynamic Cross-Variable Filtering Module was removed.
%\item \textbf{re-MLP}: The Dynamic Cross-Variable Filtering Module was replaced with a simple MLP to capture inter-variable relationships.
%\item \textbf{re-Attn}: The Dynamic Cross-Variable Filtering Module was replaced with an attention mechanism to capture inter-variable relationships.
%\end{itemize}

\begin{table}[h!]
\centering
\resizebox{0.45\textwidth}{!}{\begin{tabular}{cc|cc|cc|cc}
\toprule
\multicolumn{2}{c}{Dataset}  & \multicolumn{2}{c}{ETTm1} & \multicolumn{2}{c}{Weather} & \multicolumn{2}{c}{Electricity} \\
\cmidrule(lr){1-2} \cmidrule(lr){3-4} \cmidrule(lr){5-6} \cmidrule(lr){7-8}
\multicolumn{2}{c}{Metric} & MSE & MAE & MSE & MAE & MSE & MAE \\
\midrule
\multicolumn{2}{c}{FilterTS} & \textbf{0.385} & \textbf{0.396} & \textbf{0.244} & \textbf{0.274} & \textbf{0.180} & \textbf{0.271} \\
\midrule
\multicolumn{2}{c}{w/o-SGF} & 0.392 & 0.399 & 0.250 & 0.276 & 0.197 & 0.282 \\
\midrule
\multicolumn{2}{c}{w/o-DCF} & 0.390 & 0.400 & 0.250 & 0.278 & 0.184 & 0.274 \\
\midrule
\multicolumn{2}{c}{w/o-SGF\&DCF} & 0.405 & 0.407 & 0.267 & 0.286 & 0.208 & 0.286 \\
\midrule
\multicolumn{2}{c}{re-MLP} & 0.387 & 0.397 & 0.248 & 0.274 & 0.204 & 0.297 \\
\midrule
\multicolumn{2}{c}{re-Attn} & 0.392 & 0.398 & 0.254 & 0.281 & 0.181 & 0.273 \\
\bottomrule
\end{tabular}}
\caption{Ablation analysis of the FilterTS model, averaged across all prediction lengths for each dataset. 
}
\label{Tab2}
\end{table}

Table \ref{Tab2} shows the results of the ablation study. We observe that removing the Static Global Filtering Module results in performance degradation of 5\%, highlighting the importance of capturing dominant stable frequency components within the time series for long-term forecasting. Similarly, the removal of the Dynamic Cross-Variable Filtering Module leads to a performance drop of 2\%, indicating the benefits of effectively leveraging inter-variable relationships.

Furthermore, replacing the Dynamic Cross-Variable Filtering Module with either an MLP or an attention mechanism resulted in inferior performance. This suggests that selectively extracting shared frequency components across variables, rather than simply fusing inter-variable information, is a more effective approach for modeling inter-variable relationships in multivariate time series forecasting.

\subsection{Model Efficiency}

\begin{figure}[t!]
\centering
\includegraphics[width=1\columnwidth]{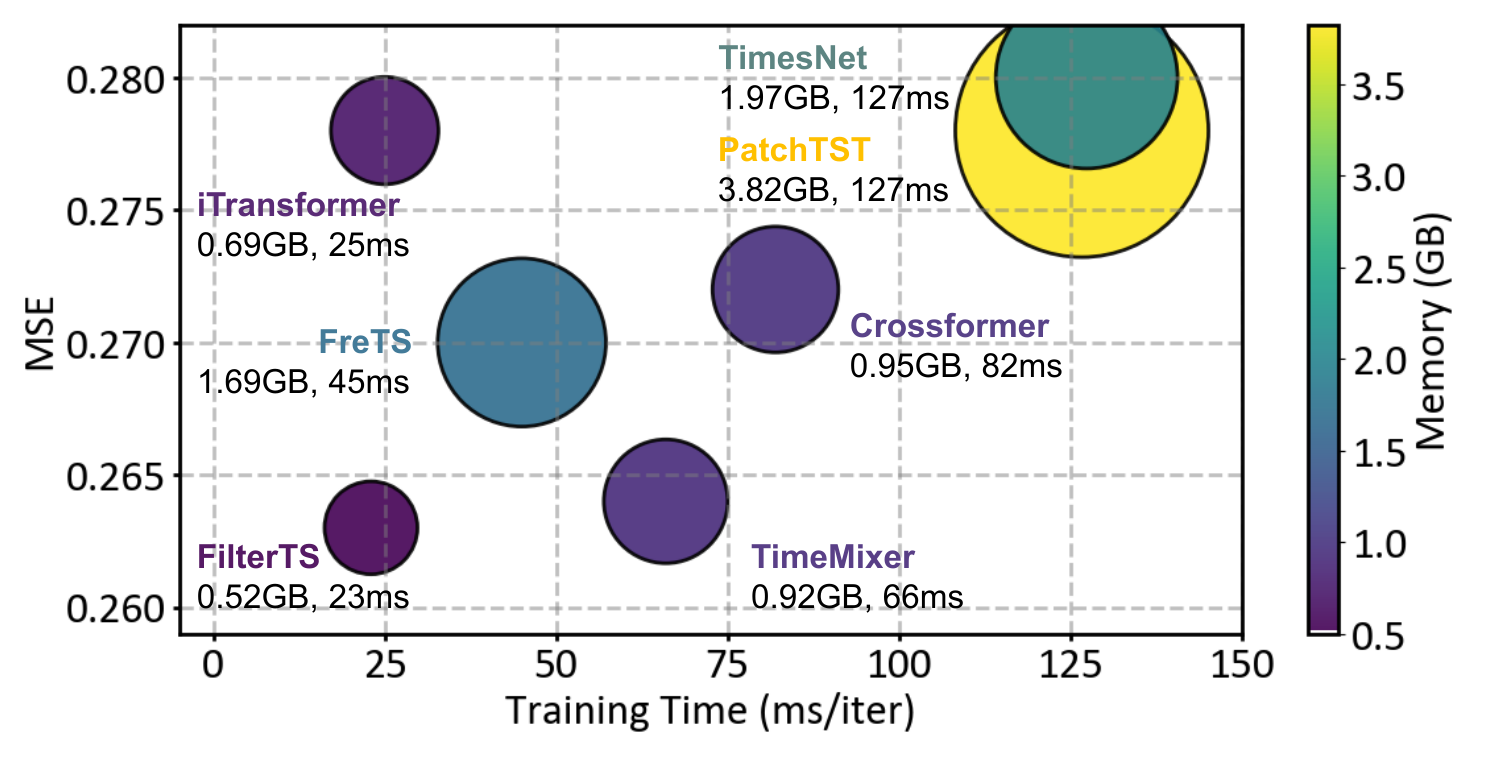} 
\caption{Performance Analysis of FilterTS: Assessing MSE, Training Time, and Memory Usage, evaluated on the Weather Dataset with a 96-In/336-Out Setup.}
\label{fig4}
\end{figure}

We compare the FilterTS model against other state-of-the-art models in terms of forecasting accuracy, memory usage, and training speed. The results, as shown in Figure \ref{fig4}, indicate that FilterTS outperforms the comparison models in predictive performance, achieves lower memory consumption, and faster training speed.

\section{Conclusion}
This paper introduced FilterTS, a novel multivariate time series forecasting model that enhances prediction accuracy and computational efficiency through comprehensive filtering techniques in the frequency domain. By incorporating both Static Global and Dynamic Cross-Variable Filtering Modules, FilterTS effectively captures essential frequency components. Our extensive experiments on eight real-world datasets demonstrate that FilterTS outperforms existing state-of-the-art methods in forecasting accuracy and efficiency. These results underscore the benefits of targeted frequency analysis in time series forecasting, suggesting that refining frequency components can significantly advance predictive capabilities.

\section{Acknowledgments}
This work was supported in part by the Natural Science Foundation of Tianjin of China under Grant No. 21JCZDJC00740.

\bibliography{aaai25}

\clearpage

\appendix
\section*{\centering{APPENDIX}}

% ---------- 重新编号 ----------
\setcounter{figure}{0}
\renewcommand{\thefigure}{A\arabic{figure}}   % 图：A1, A2, …
\setcounter{table}{0}
\renewcommand{\thetable}{A\arabic{table}}     % 表：A1, A2, …
\setcounter{equation}{0}
\renewcommand{\theequation}{A\arabic{equation}} % 公式：A1, A2, …

\section{Efficient Linear Convolution Using Fast Fourier Transform}

In the main text, we discussed how the Fast Fourier Transform (FFT) is applied to normalized time series data after zero-padding to efficiently transform these series into the frequency domain. This extension is crucial as it ensures that the multiplication in the frequency domain accurately represents linear convolution in the time domain, rather than circular convolution. 

To understand this, a distinction must be drawn between linear and circular convolution. Linear convolution, which involves the sliding of one signal over another to produce a third signal, is essential when using time series data as filters in signal processing applications. It provides a complete response incorporating all shifts of the input signals, unlike circular convolution which inherently assumes periodic signal continuation and can lead to edge artifacts when such assumptions are invalid.
By padding zeros, we extend the discrete-time sequences sufficiently to eliminate the circular nature of convolution performed via FFT, thus mimicking the linear convolution operation without computational inefficiency.

Next, we will present a proof that the frequency domain multiplication of the zero-padded sequences after their Fourier transformation is equivalent to the linear convolution of the original sequences in the time domain, followed by an analysis of the computational efficiency of frequency domain multiplication versus time domain linear convolution.

\subsection{Theoretical Proof}
\textbf{Theorem} Let \(x(n)\) and \(h(n)\) be discrete-time sequences of respective lengths \(N\) and \(M\). Zero-pad \(x(n)\) and \(h(n)\) to form sequences \(\tilde{x}(n)\) and \(\tilde{h}(n)\), each of length \(L \geq N + M - 1\). The linear convolution \(y(n)\) of \(x(n)\) and \(h(n)\) is defined as:
\begin{equation}
\begin{split}
y(n) &= (x \ast h)(n) \\
&= \sum_{k=0}^{N-1} x(k) h(n-k), \quad \text{for } 0 \leq n < N+M-1.
\end{split}
\end{equation}
Perform the Fast Fourier Transform (FFT) on the zero-padded sequences \(\tilde{x}(n)\) and \(\tilde{h}(n)\), resulting in \(\mathcal{X}(f)\) and \(\mathcal{H}(f)\). The product of these two frequency domain representations \(\mathcal{Y}(f)\) represents the frequency domain form of \(y(n)\), i.e.:
\begin{equation}
\mathcal{Y}(f) = \mathcal{X}(f) \cdot \mathcal{H}(f),
\end{equation}
where \(\mathcal{Y}(f)\) is the Fourier transform of the zero-padded convolution result.

\textbf{Proof.}
Given the zero-padded sequences \(\tilde{x}(n)\) and \(\tilde{h}(n)\), the Fast Fourier Transform (FFT) of \(\tilde{x}(n)\) and \(\tilde{h}(n)\) respectively yield:
\begin{equation}
\mathcal{X}(f) = \sum_{n=0}^{L-1}\tilde{x}(n)e^{-i2\pi \frac{fn}{L}}, \quad \mathcal{H}(f) = \sum_{n=0}^{L-1}\tilde{h}(n)e^{-i2\pi \frac{fn}{L}}.
\end{equation}
The product \(\mathcal{X}(f) \cdot \mathcal{H}(f)\) then becomes:
\begin{equation}
\mathcal{X}(f) \cdot \mathcal{H}(f) = \left(\sum_{n=0}^{L-1}\tilde{x}(n)e^{-i2\pi \frac{fn}{L}}\right) \cdot \left(\sum_{m=0}^{L-1}\tilde{h}(m)e^{-i2\pi \frac{fm}{L}}\right),
\end{equation}
applying the definition of convolution in the frequency domain:
\begin{equation}
= \sum_{n=0}^{L-1} \sum_{m=0}^{L-1} \tilde{x}(n)\tilde{h}(m)e^{-i2\pi \frac{f(n+m)}{L}}.
\end{equation}
Substituting \(p = n + m\) and adjusting the limits appropriately, we get:
\begin{equation}
= \sum_{n=0}^{L-1}\sum_{p=n}^{2L-2}   \tilde{x}(n)\tilde{h}(p-n) e^{-i2\pi \frac{fp}{L}},
\end{equation}
which, due to the zero-padding, simplifies within the limits as \(\tilde{x}(n) = 0\) for \(n \geq N\) and \(\tilde{h}(m) = 0\) for \(m \geq M\) or \(m < 0\), ensuring that the above expression is equivalent to:
\begin{equation}
\begin{split}
&= \sum_{n=0}^{N-1}\sum_{p=0}^{L-1}   \tilde{x}(n)\tilde{h}(p-n) e^{-i2\pi \frac{fp}{L}}\\
&= \sum_{p=0}^{L-1}\sum_{n=0}^{N-1}   \tilde{x}(n)\tilde{h}(p-n) e^{-i2\pi \frac{fp}{L}}
\end{split}
\end{equation}
Substituting  \(k = n\) and \(n = p\)  for consistency, we get:

\begin{equation}
\begin{split}
&= \sum_{n=0}^{L-1}\left(\sum_{k=0}^{N-1}   \tilde{x}(k)\tilde{h}(n-k)\right) e^{-i2\pi \frac{fn}{L}}\\
&= \sum_{n=0}^{L-1} (\tilde{x} \ast \tilde{h})(n) e^{-i2\pi \frac{fn}{L}} = \mathcal{Y}(f)
\end{split}
\end{equation}
The inverse FFT of \(\mathcal{Y}(f)\) then recovers \(\tilde{y}(n)\):
\begin{equation}
\tilde{y}(n) = \frac{1}{L} \sum_{f=0}^{L-1} \mathcal{Y}(f) e^{i2\pi \frac{fn}{L}}.
\end{equation}
By the properties of the FFT and convolution, \(\tilde{y}(n)\) in the time domain represents the circular convolution of \(\tilde{x}(n)\) and \(\tilde{h}(n)\), which, due to the zero-padding, is identical to the linear convolution \(y(n)\) of \(x(n)\) and \(h(n)\) for \(n < N+M-1\). Therefore, we have:
\begin{equation}
y(n) = \tilde{y}(n) = \sum_{f=0}^{L-1} \left( \mathcal{X}(f) \cdot \mathcal{H}(f) \right) e^{i2\pi \frac{fn}{L}} \cdot \frac{1}{L},
\end{equation}
and thus we have shown that the Fast Fourier Transform of the zero-padded sequences \(\mathcal{Y}(f) = \mathcal{X}(f) \cdot \mathcal{H}(f) \) corresponds exactly to the Fourier transform of the linear convolution of the original sequences \(x(n)\) and \(h(n)\), thereby completing the proof that \(\mathcal{Y}(f) = \mathcal{X}(f) \cdot \mathcal{H}(f) \) accurately represents the linear convolution \(y(n) = (x \ast h)(n)\) under zero-padding and sufficient length \(L\). Therefore, the theorem is proved.
\hfill $\square$

\subsection{Computational Efficiency}

The application of the Fast Fourier Transform (FFT) in linear convolution significantly enhances computational efficiency for handling large-scale data in digital signal processing. By transforming time-domain data into the frequency domain, convolution operations are executed as point-wise multiplications followed by an inverse FFT. This transition changes the computational complexity from \(O(N^2)\), typical for traditional time-domain convolution with sequences of length \(N\), to \(O(N \log N)\) when considering zero-padded sequences to prevent circular convolution effects.

In traditional time-domain convolution, the computational effort involves calculating the convolution sum for each shift of the overlapping sequences, resulting in \(N^2\) operations if the sequences are of length \(N\). This approach becomes increasingly untenable for large \(N\) as it necessitates \(N\) multiplications and \(N-1\) additions for each of the \(N+M-1\) output values, culminating in a complexity of \(O((N+M-1) \times N)\). When \(N\) is approximately equal to \(M\), this complexity simplifies to \(O(N^2)\).

Conversely, in frequency-domain convolution via FFT, the sequences are first zero-padded to a minimum length of \(L = N + M - 1\), typically approximating \(2N\) when \(N \approx M\). After computing the FFT of each zero-padded sequence, point-wise multiplication is performed, followed by an inverse FFT. The computational complexity for these FFT operations is \(O(L \log L)\), which simplifies to \(O(2N \log (2N))\) under typical conditions where \(L \approx 2N\). Using properties of logarithms, this can be expressed as \(O(2N \log 2 + 2N \log N)\), which further simplifies to \(O(N \log N)\) given the constant factors.

%Thus, comparing the computational complexities, traditional time-domain convolution has a complexity of \(O(N^2)\), which becomes prohibitive with increasing \(N\). In contrast, FFT-based convolution reduces the complexity to \(O(N \log N)\), offering substantial computational savings and making it feasible and efficient for large data sets. This stark contrast underscores the advantages of using FFT for convolution tasks in digital signal processing, where efficiency and speed are paramount.

\section{Complex-Valued Operations in FilterTS}
This section outlines the essential mathematical formulations of the complex-valued operations used in our model, designed to operate directly with complex numbers. Here, we provide the mathematical definitions of complex-valued operations that are not yet implemented in PyTorch. The definitions for the Complex Linear Layer and Complex ReLU, draw upon the formulations found in the works of \cite{yi2024frequency} and \cite{trabelsi2018deep}, respectively.

\subsection{Complex Linear Layer}
The complex linear layer employs a transformation designed to handle the separate real and imaginary parts of complex-valued input data. The operations are governed by the following equations, which ensure the integrity of complex number properties:

\begin{equation}
    \text{Re}(\mathbf{X}_{\text{out}}) = \mathbf{W}_{\text{real}} \cdot \text{Re}(\mathbf{X}_{\text{mix}}) - \mathbf{W}_{\text{imag}} \cdot \text{Im}(\mathbf{X}_{\text{mix}})
\end{equation}
\begin{equation}
    \text{Im}(\mathbf{X}_{\text{out}}) = \mathbf{W}_{\text{imag}} \cdot \text{Re}(\mathbf{X}_{\text{mix}}) + \mathbf{W}_{\text{real}} \cdot \text{Im}(\mathbf{X}_{\text{mix}})
\end{equation}

Here, \( \mathbf{W}_{\text{real}} \) and \( \mathbf{W}_{\text{imag}} \) are the weight matrices for the real and imaginary components, respectively. This dual-path transformation ensures that both amplitude and phase information are preserved, crucial for the faithful representation of signals in the frequency domain.

\subsection{Complex LayerNorm}
Complex LayerNorm is designed to stabilize the training of neural networks by independently normalizing the real and imaginary components of complex-valued inputs. This normalization is crucial for reducing internal covariate shift, thereby facilitating faster and more stable convergence. The operation is mathematically expressed as follows:

Assuming \( \mathbf{x} = \text{Re}(\mathbf{x}) + i \text{Im}(\mathbf{x}) \), where \( \text{Re}(\mathbf{x}) \) and \( \text{Im}(\mathbf{x}) \) represent the real and imaginary components of \( \mathbf{x} \), respectively, the normalization for each component is computed independently:

\begin{equation}
    \text{Re}(\text{CLayerNorm}(\mathbf{x})) = \text{LayerNorm}(\text{Re}(\mathbf{x}))
\end{equation}
\begin{equation}
    \text{Im}(\text{CLayerNorm}(\mathbf{x})) = \text{LayerNorm}(\text{Im}(\mathbf{x}))
\end{equation}

This approach ensures that both the real and imaginary parts of the complex-valued input are treated separately yet consistently, promoting the integrity and efficiency of learning processes involving complex-valued data.

\subsection{Complex ReLU}
The Complex ReLU activation function is designed to handle the non-linear activation of complex numbers by independently applying the ReLU function to both the real and imaginary components of each neuron's output. This method ensures that the inherent complex structure is respected during the activation process. Mathematically, this can be represented as:

Let \( z = \text{Re}(z) + i\text{Im}(z) \), where \( \text{Re}(z) \) and \( \text{Im}(z) \) are the real and imaginary parts of \( z \), respectively. The CReLU function is applied as follows:

\begin{equation}
    \text{Re}(\text{CReLU}(z)) = \text{ReLU}(\text{Re}(z))
\end{equation}
\begin{equation}
    \text{Im}(\text{CReLU}(z)) = \text{ReLU}(\text{Im}(z))
\end{equation}

This formulation ensures that each part of the complex-valued input is treated separately, allowing for phase preservation and appropriate nonlinear transformation in networks dealing with complex-valued data.

\subsection{Complex Softmax}
The Complex Softmax function applies the softmax function to the magnitudes while preserving the phases of the complex numbers. This formulation is as follows:

Let \( z = \text{Re}(z) + i\text{Im}(z) \).

Compute the softmax over the magnitudes:
\begin{equation}
    S_m = \frac{\exp(\sqrt{\text{Re}(z)^2 + \text{Im}(z)^2})}{\sum \exp(\sqrt{\text{Re}(z)^2 + \text{Im}(z)^2})}
\end{equation}

Define the phase factor \( \phi \) by normalizing \( z \):
\begin{equation}
    \phi = \frac{z}{|z|} = \frac{\text{Re}(z) + i\text{Im}(z)}{\sqrt{\text{Re}(z)^2 + \text{Im}(z)^2}}
\end{equation}

Decompose \( \phi \) into its real and imaginary parts:
\begin{equation}
    \text{Re}(\phi) = \frac{\text{Re}(z)}{\sqrt{\text{Re}(z)^2 + \text{Im}(z)^2}}
\end{equation}
\begin{equation}
    \text{Im}(\phi) = \frac{\text{Im}(z)}{\sqrt{\text{Re}(z)^2 + \text{Im}(z)^2}}
\end{equation}

Finally, express the Complex Softmax as:
\begin{equation}
    \text{Re}(\text{CSoftmax}(z)) = S_m \cdot \text{Re}(\phi)
\end{equation}
\begin{equation}
    \text{Im}(\text{CSoftmax}(z)) = S_m \cdot \text{Im}(\phi)
\end{equation}

This method ensures that the softmax is applied to the magnitudes of the complex numbers, and the phases, represented by \( \phi \), are meticulously preserved.

\section{Experimental Details}
\subsection{Dataset Descriptions}
We evaluate our proposed models on eight multivariate time series datasets, each representing different domains such as energy, economy, transportation, and climatology, to ensure the robustness and applicability of our findings. Here we provide detailed descriptions of each dataset:

\begin{enumerate}
    \item \textbf{ETT}: The ETT datasets, namely ETTh1, ETTh2, ETTm1, and ETTm2, consist of measurements from electrical transformers. ETTh datasets (ETTh1, ETTh2) record seven variables including voltage, current, and temperature on an hourly basis, while ETTm datasets (ETTm1, ETTm2) capture the same seven variables every 15 minutes, from July 2016 to July 2018.

    \item \textbf{Electricity}: This dataset tracks the electricity consumption metrics of 321 clients, recorded every 15 minutes, reflecting residential and industrial usage. It involves a large number of variables, with 321 distinctive measures of consumption patterns.

    \item \textbf{Exchange Rate}: Featuring daily records of exchange rates for eight major currencies, this dataset encompasses a time span from 1990 to 2016 and includes eight variables per timestamp, aiding in the analysis of economic trends.

    \item \textbf{Traffic}: Capturing the dynamics of traffic flow and occupancy rates with 862 sensors, this dataset provides hourly data across various freeways in the San Francisco Bay Area from January 2015 to December 2016. The dataset is rich in dimensions, focusing on a broad range of traffic-related variables.

    \item \textbf{Weather}: This dataset is gathered every 10 minutes from the Max Planck Institute for Biogeochemistry’s weather station and includes a comprehensive 21 meteorological variables such as temperature, humidity, and wind speed throughout 2020. It offers a detailed look into climatic conditions with a high resolution in time and variable space.
\end{enumerate}

In our study, we adopt the protocol for data processing and splitting into training, validation, and testing sets as established by iTransformer \cite{liu2024itransformer}. Further details on the datasets can be found in Table  \ref{tab1}.

\begin{table}[h!]
    \centering
    \caption{Detailed dataset descriptions. \textit{Dim} denotes the variate number of each dataset. \textit{Split} represents the proportion of data points allocated to the training, validation, and testing sets, respectively. \textit{Frequency} denotes the sampling interval of time points.}
    \label{tab:dataset_descriptions}
    \begin{tabular}{lllll}
        \toprule
        Dataset & Dim & Split& Frequency & Domain \\
        \midrule
        ETTm1 & 7 & 6 : 2 : 2 & 15min & Electricity\\
        ETTm2 & 7 & 6 : 2 : 2 & 15min & Electricity\\
        ETTh1 & 7 & 6 : 2 : 2 & 1hour & Electricity\\
        ETTh2 & 7 & 6 : 2 : 2 & 1hour & Electricity\\ 
        ECL & 321 & 7 : 1 : 2 & 1hour & Electricity\\
        Exchange & 8 & 7 : 1 : 2 & 1day & Economy\\
        Traffic & 862 &  7 : 1 : 2  & 1hour & Transportation\\
        Weather & 21 &  7 : 1 : 2 & 10min & Climatology\\
        \bottomrule
    \end{tabular}
    \label{tab1}
\end{table}

%\subsection{Baseline Descriptions}

%In this paper, we introduced FilterTS, a novel multivariate time series forecasting model that leverages frequency-domain analysis and advanced filtering techniques to enhance prediction accuracy and computational efficiency. FilterTS incorporates both Static Global Filtering Modules and Dynamic Cross-Variable Filtering Modules to effectively capture stable and variable frequency components. Our extensive experiments on eight real-world datasets demonstrate that FilterTS outperforms existing state-of-the-art methods in forecasting accuracy and efficiency. These results underscore the transformative potential of employing comprehensive filtering techniques to refine frequency representations, thereby advancing the field of time series forecasting.
\begin{figure*}[h!]
    \centering

    \begin{subfigure}[b]{0.48\textwidth} 
        \centering
        \includegraphics[width=\textwidth]{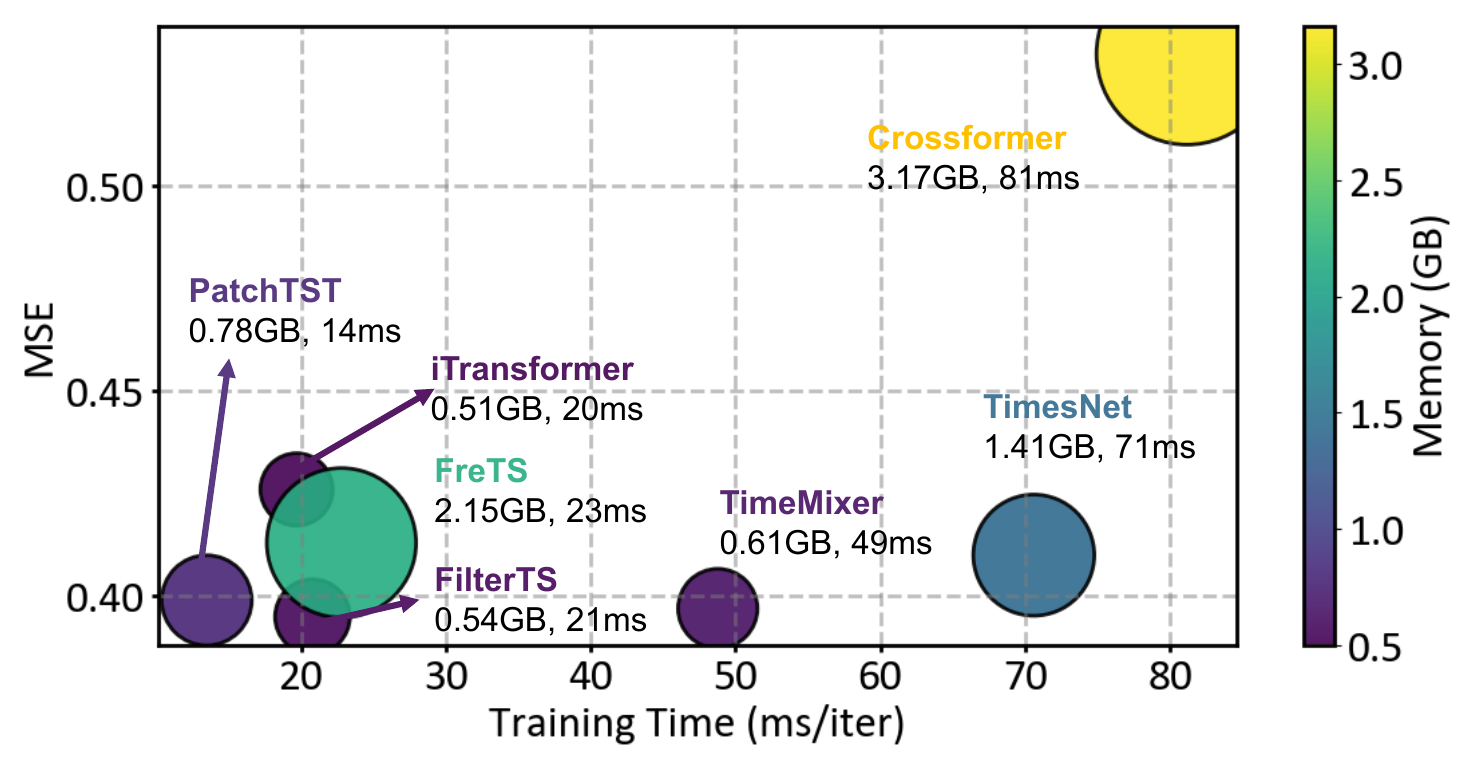} 
        \caption{ETTm1} 
        \label{ETTm1} 
    \end{subfigure}
    \hfill
    \begin{subfigure}[b]{0.48\textwidth} 
        \centering
        \includegraphics[width=\textwidth]{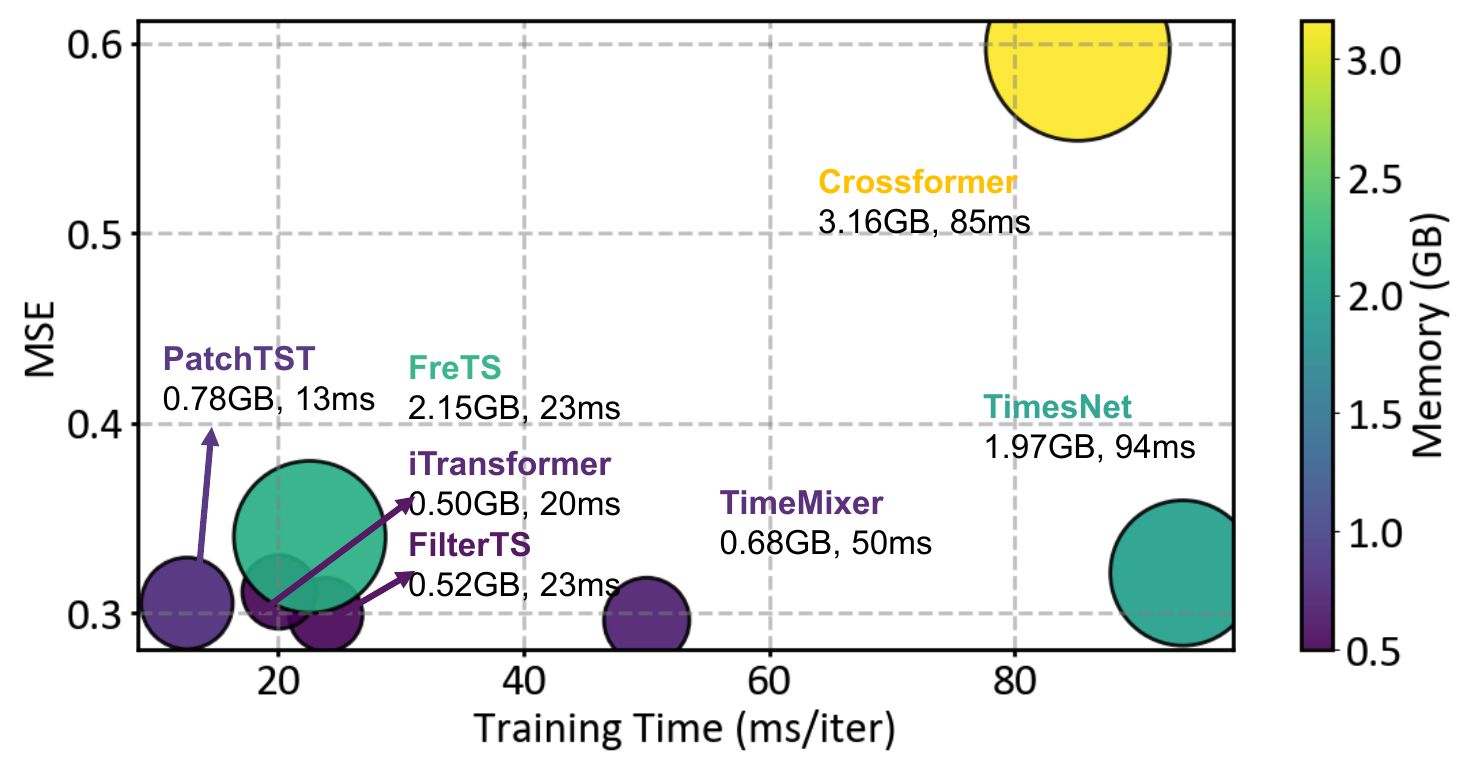} 
        \caption{ETTm2} 
        \label{ETTm2} 
    \end{subfigure}

    %\vspace{0.3cm} % 添加垂直间距

    \begin{subfigure}[b]{0.48\textwidth} 
        \centering
        \includegraphics[width=\textwidth]{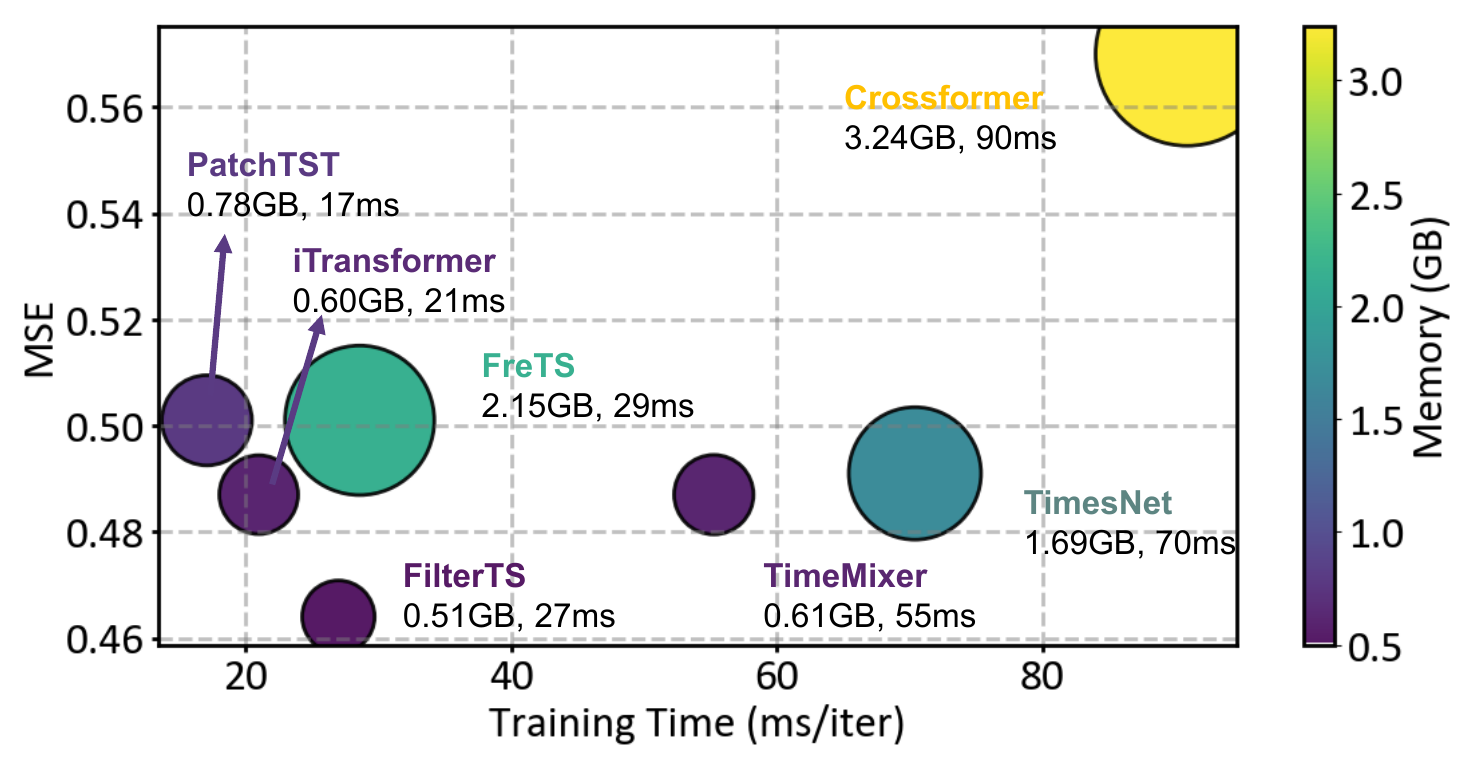} 
        \caption{ETTh1} 
        \label{ETTh1} 
    \end{subfigure}
    \hfill
    \begin{subfigure}[b]{0.48\textwidth} 
        \centering
        \includegraphics[width=\textwidth]{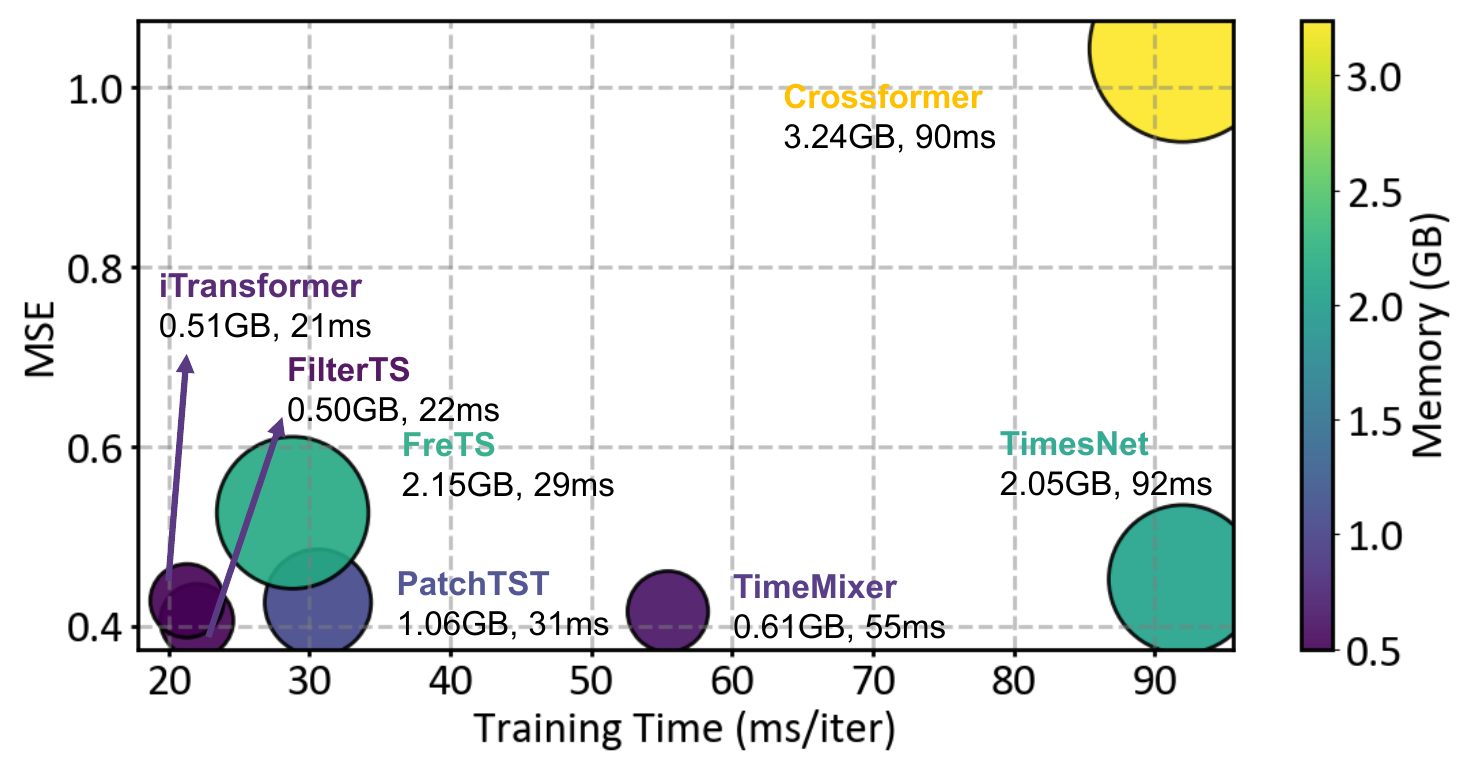} 
        \caption{ETTh2} 
        \label{ETTh2} 
    \end{subfigure}

    %\vspace{0.3cm} % 添加垂直间距

    \begin{subfigure}[b]{0.48\textwidth} 
        \centering
        \includegraphics[width=\textwidth]{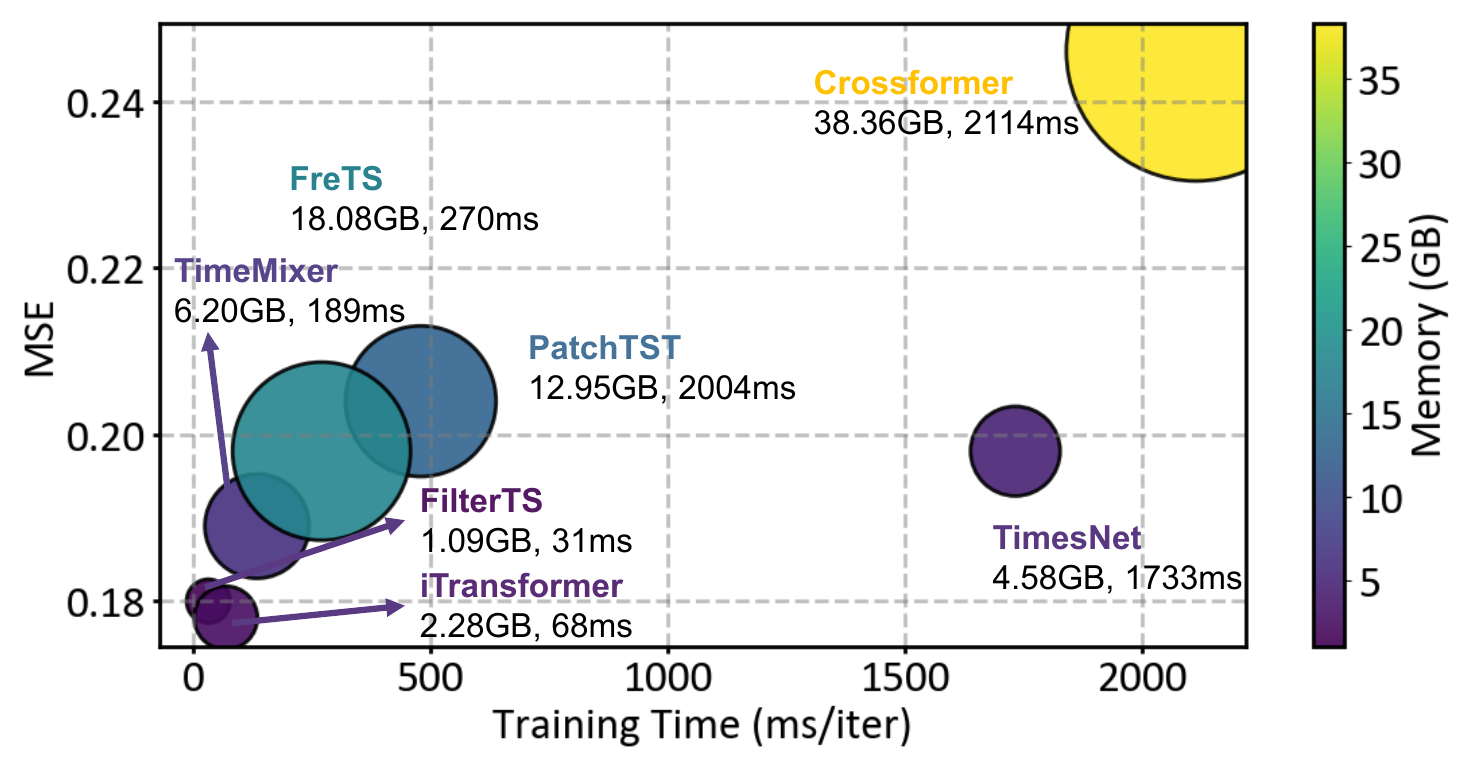} 
        \caption{Electricity} 
        \label{Electricity} 
    \end{subfigure}
    \hfill
    \begin{subfigure}[b]{0.48\textwidth} 
        \centering
        \includegraphics[width=\textwidth]{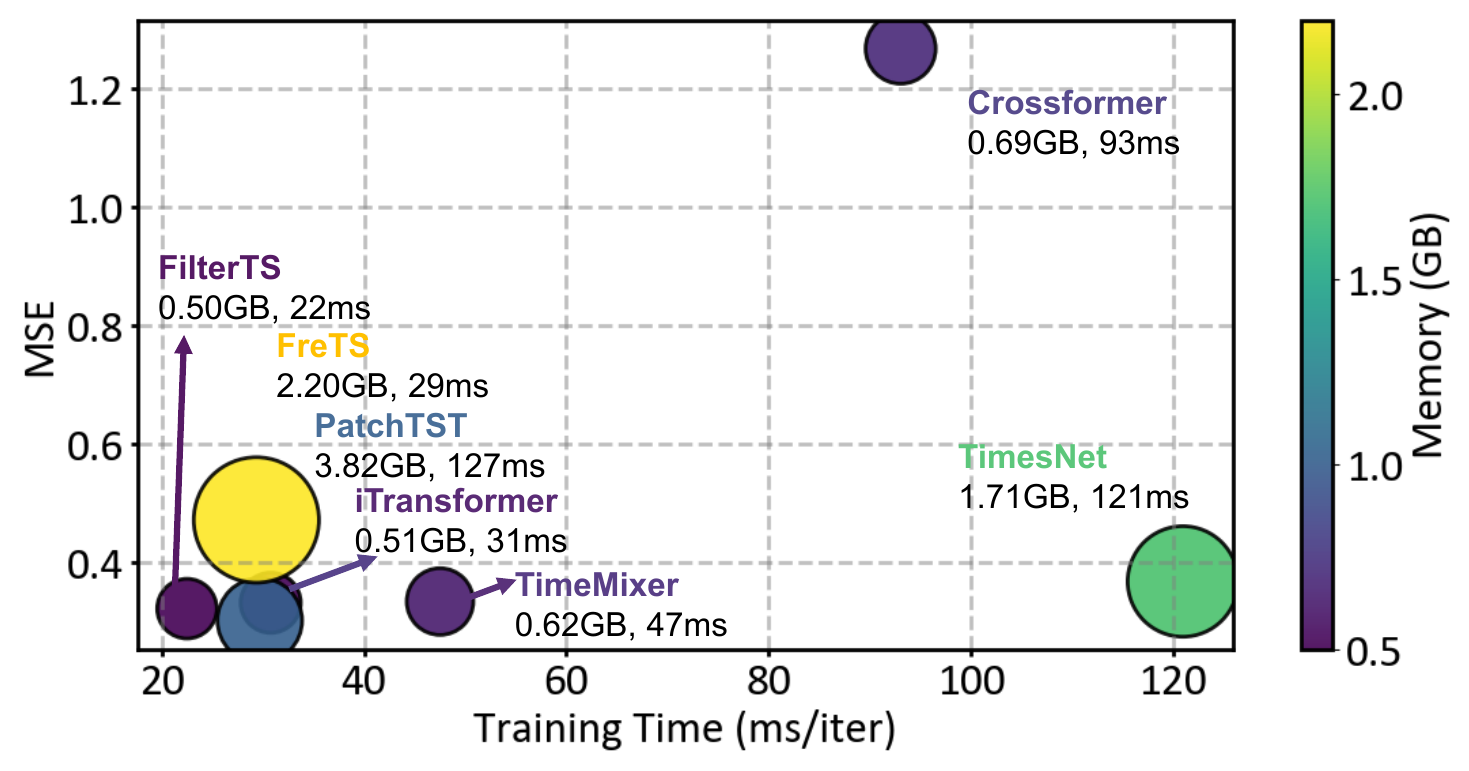} 
        \caption{Exchange} 
        \label{Exchange} 
    \end{subfigure}

    %\vspace{0cm} % 添加垂直间距

    \begin{subfigure}[b]{0.48\textwidth} 
        \centering
        \includegraphics[width=\textwidth]{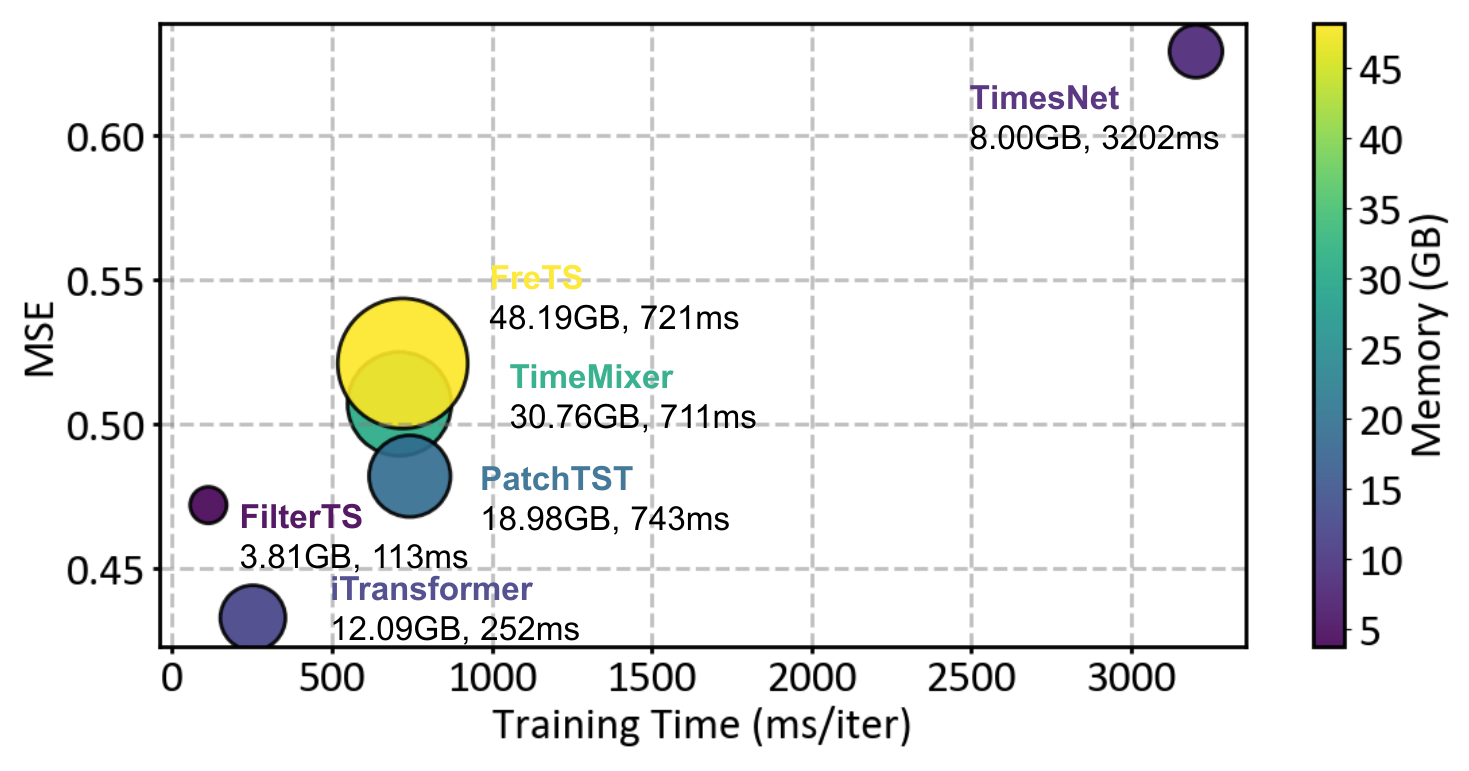} 
        \caption{Traffic} 
        \label{Traffic} 
    \end{subfigure}
    \hfill
    \begin{subfigure}[b]{0.48\textwidth} 
        \centering
        \includegraphics[width=\textwidth]{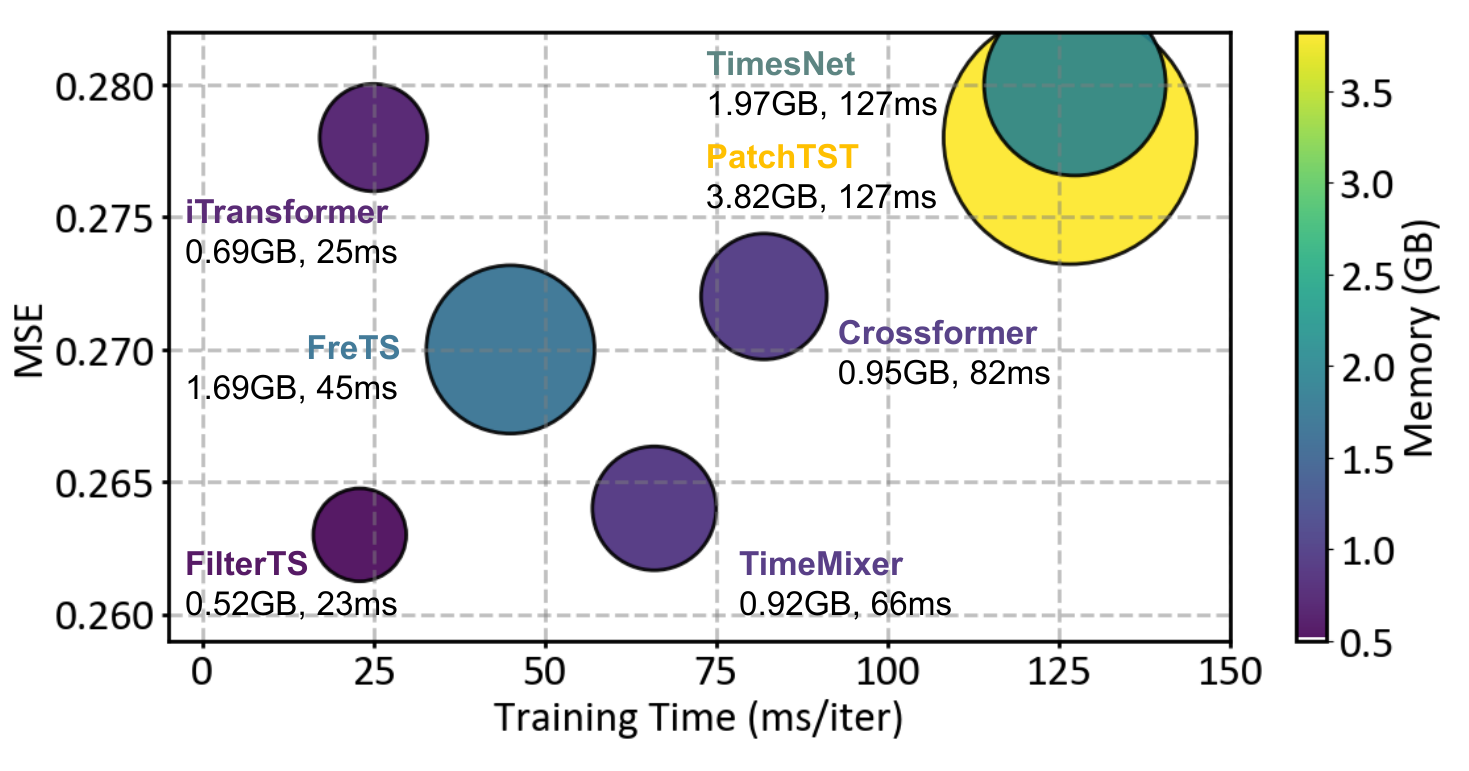} 
        \caption{Weather} 
        \label{Weather} 
    \end{subfigure}

    \caption{Comprehensive Performance Analysis of FilterTS: Assessing MSE, Average Training Time per Iteration during the First Epoch, and Memory Usage (lower is better) at Batch Size 32, evaluated across Various Datasets with a 96-In/336-Out Setup.  }
    \label{mainfigure} 
\end{figure*}

\begin{figure*}[t]
    \centering

    \begin{subfigure}[b]{0.33\textwidth} 
        \centering
        \includegraphics[width=\textwidth]{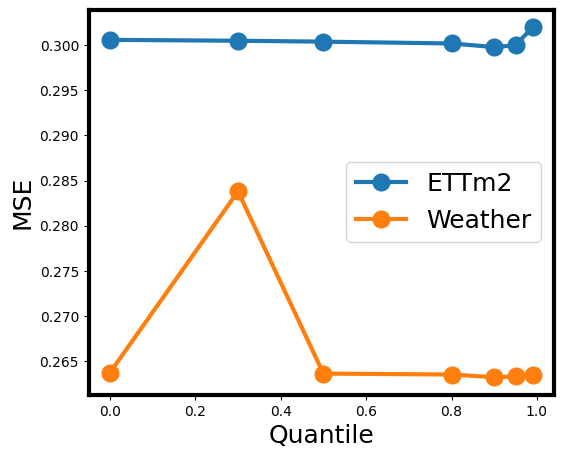} 
        \caption{}%$\alpha-$Quantile} 
        \label{Quantile} 
    \end{subfigure}
    \hfill
    \begin{subfigure}[b]{0.33\textwidth} 
        \centering
        \includegraphics[width=\textwidth]{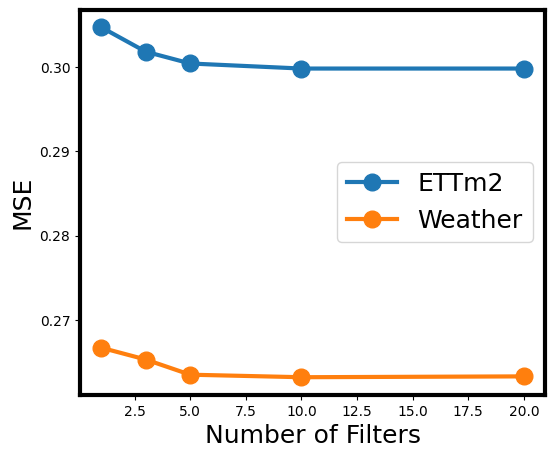} 
        \caption{}%Number of Static Filters} 
        \label{Number} 
    \end{subfigure}
    \hfill
    \begin{subfigure}[b]{0.33\textwidth} 
        \centering
        \includegraphics[width=\textwidth]{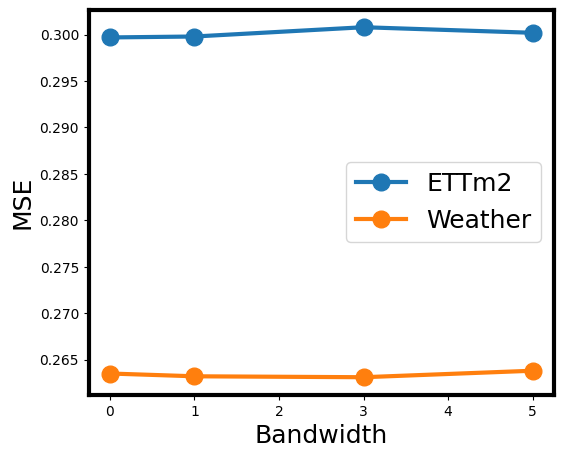} 
        \caption{}%Bandwidth} 
        \label{BandWidth} 
    \end{subfigure}

\caption{Hyper-Parameter Sensitivity Analysis of the FilterTS Model on the ETTm2 and Weather Datasets with a 96-In/336-Out Setup: (a) Impact of the $\alpha$-Quantile on model performance; (b) Effects of varying the number of global static filters (\textit{num\_static\_filters}); (c) Influence of global static filter bandwidth (\textit{delta\_bandwidth}).}

    \label{hyper} 
\end{figure*}
\subsection{Implementation Details}
Experiments were executed using the Python 3.10 environment, with PyTorch version 2.2.0 and CUDA 11.8, on a system featuring an NVIDIA RTX 4090 GPU equipped with 24GB of memory. The experimental framework was informed by the configuration used for iTransformer\cite{liu2024itransformer}, setting the lookback window to 96 timesteps and extending prediction horizons to \{96, 192, 336, 720\}. 
Optimization was facilitated using the Adam optimizer, with the learning rate being halved after each training epoch. The Mean Squared Error (MSE) served as the loss metric. Consistency was maintained across experiments with a fixed batch size of 32 and a total of 10 training epochs.
A systematic grid search was utilized to optimize the initial learning rate, exploring values within the set \{0.0001, 0.0005, 0.001, 0.005\}. The architecture's complexity was adjusted through variations in hidden layer sizes \{128, 256, 512\} and the number of model layers ranging from 1 to 4. 
For FilterTS, the threshold for significant frequency components (\textit{quantile}) was set to 0.9. The number of band-pass filters (\textit{num\_static\_filters}) was set to 10, each with a bandwidth (\textit{delta\_bandwidth}) of 1.

\section{Further Analysis}
\subsection{Full Performance Analysis}
In Figure \ref{mainfigure}, we present a comparative analysis of the FilterTS model against other state-of-the-art models, focusing on forecasting accuracy, memory usage, and training speed. The results indicate that FilterTS outperforms comparison models in predictive performance in the majority of cases, and achieves lower memory consumption and faster training speeds, affirming the efficacy of our model's design. Particularly notable is the performance on datasets with a high number of variables, such as Weather with 21 variables, Electricity with 321 variables, and Traffic with 862 variables. FilterTS demonstrates significantly lower memory requirements; for example, on the Traffic dataset, it requires only 3.81GB of memory compared to 12.09GB for iTransformer, 18.98GB for PatchTST, and a substantial 30.76GB for TimeMixer. This stark contrast highlights FilterTS's exceptional suitability for multivariate time series forecasting tasks involving extensive variable counts, showcasing substantial advantages in operational efficiency and making it an ideal choice for complex forecasting scenarios.

\subsection{Hyper-Parameter Sensitivity}

In this section, we analyze the sensitivity of key hyperparameters in the FilterTS model. We assess how these parameters influence the model's performance on the ETTm2 and Weather datasets. The results of these analyses are illustrated in Figure \ref{hyper}, which depicts the impact of each parameter on the Mean Squared Error (MSE).

\textbf{Quantile in Dynamic Cross-Variable Filtering Module:}
The dynamic filters in FilterTS are critical for emphasizing significant frequency components while minimizing noise interference. The selection of the quantile threshold (\textit{quantile}) plays a pivotal role in this process. As shown in Figure A2a, the MSE decreases with an increase in the \textit{quantile} up to a point, after which it begins to increase, peaking at a quantile of 0.9. This pattern suggests that lower quantiles may allow noise to affect the model, whereas excessively high quantiles might exclude vital frequency components. Consequently, a quantile of 0.9 is optimal, balancing noise reduction and information retention.

\textbf{Number of Global Static Filters:}
Figure A2b examines the effect of varying the number of static filters (\textit{num\_static\_filters}). Increasing the number of filters initially decreases the MSE, indicating enhanced capture of inherent frequency components. However, beyond a certain point, the benefits plateau, implying that too many filters might capture extraneous frequencies, leading to overfitting. We have chosen to set the number of static filters at ten, which provides a balanced approach to capturing the essential frequencies without introducing unnecessary complexity.

\textbf{Global Static Filter Bandwidth:}
The bandwidth of the static filters (\textit{delta\_bandwidth}) also affects model performance, as illustrated in Figure A2c. A narrow bandwidth can precisely capture relevant frequencies but may miss slight shifts in the frequency domain. Conversely, a wider bandwidth, while accommodating frequency shifts, introduces more noise into the model. Our analysis shows that a bandwidth of 1 strikes an effective balance by accommodating potential frequency deviations without significant noise increase, thus optimizing the model’s performance.

These findings underscore the importance of carefully tuning the hyperparameters in FilterTS to maximize forecasting accuracy while maintaining computational efficiency. The selected settings ensure that FilterTS is robust across various datasets with differing characteristics, particularly those with large numbers of variables where memory efficiency and processing speed are crucial.

%\subsection{Model Stability Analysis}
%To ascertain the stability of our FilterTS model, we conducted experiments using three distinct random seeds and reported the variance in Mean Squared Error (MSE) to evaluate performance consistency. The outcomes, as presented in Table 2, confirm that the model exhibits robust stability across different initializations. This demonstrates FilterTS's reliability in producing repeatable and reliable forecasting results, underscoring its suitability for deployment in diverse operational environments where consistency is crucial.

\end{document}